\documentclass[10pt,twocolumn,letterpaper]{article}

\usepackage{cvpr}
\usepackage{times}
\usepackage{epsfig}
\usepackage{graphicx}
\usepackage{amsmath}
\usepackage{amssymb}
\usepackage{subfigure}
\usepackage{booktabs}
\usepackage[table,xcdraw]{xcolor}
\usepackage{float}
\usepackage{stfloats}
\usepackage{enumitem}
\usepackage[ruled]{algorithm2e}
\usepackage{booktabs}
\usepackage{multirow}
\usepackage{graphicx}
\usepackage{authblk}
\cvprfinalcopy

\usepackage[pagebackref=true,breaklinks=true,letterpaper=true,colorlinks,urlcolor=blue,linkcolor=blue,citecolor=blue,bookmarks=false]{hyperref}

\newcommand{\red}[1]{\textcolor[rgb]{1,0,0}{#1}}

\def\cvprPaperID{5924} 

\ifcvprfinal\fi
\begin{document}

\title{Spatial Attentive Single-Image Deraining with a High Quality Real Rain Dataset
\vspace{-0.1in}}

\author{Tianyu Wang$^{1,2*}$ \ \ \ \ Xin Yang$^{1,2*}$ \ \ \ \ Ke Xu$^{1,2}$ \ \ \ \ Shaozhe Chen$^{1}$ \ \ \ \ Qiang Zhang$^{1}$ \ \ \ \ Rynson~W.H.~Lau$^{2\dagger}$ \\
$^{1}$Dalian University of Technology \ \ \ \ $^{2}$City University of Hong Kong\\

\text{Project Page: \url{https://stevewongv.github.io/derain-project.html}}

\vspace{-0.2in}
}


%

\maketitle
\thispagestyle{empty}

\begin{abstract}
Removing rain streaks from a single image has been drawing considerable attention as rain streaks can severely degrade the image quality and affect the performance of existing outdoor vision tasks.
%
While recent CNN-based derainers have reported promising performances, deraining remains an open problem for two reasons. First, existing synthesized rain datasets have only limited realism, in terms of modeling real rain characteristics such as rain shape, direction and intensity.
Second, there are no public benchmarks for quantitative comparisons on real rain images, which makes the current evaluation less objective. The core challenge is that real world rain/clean image pairs cannot be captured at the same time.
In this paper, we address the single image rain removal problem in two ways.
First, we propose a semi-automatic method that incorporates temporal priors and human supervision to generate a high-quality clean image from each input sequence of real rain images. Using this method,
we construct a large-scale dataset of $\sim$$29.5K$ rain/rain-free image pairs that covers a wide range of natural rain scenes.
%
Second, to better cover the stochastic distribution of real rain streaks, we propose a novel SPatial Attentive Network (SPANet) to remove rain streaks in a local-to-global manner.
Extensive experiments demonstrate that our network performs favorably against the state-of-the-art deraining methods.
\vspace{-0.1in}

\end{abstract}
\section{Introduction}
Images taken under various rain conditions often show low visibility, which can significantly affect the performance of some outdoor vision tasks, e.g., pedestrian detection~\cite{mao:cvpr:2017:pd}, visual tracking~\cite{song:cvpr:2018:vital}, or road sign recognition~\cite{zhu:cvpr:2016:tsd}. Hence, removing rain streaks from input rain images is an important research problem. In this paper, we focus on the single-image rain removal problem.{\let\thefootnote\relax\footnote{{$^{*}$ Joint first authors. $^{\dagger}$ Rynson Lau is the corresponding author, and he led this project.}}}

In the last decade, we have witnessed a continuous progress on rain removal research with many methods proposed~\cite{kang:tip:2012:imgdecomp,luo:iccv:2015:dsc,li:cvpr:2016:lp,chang:iccv:2017:lpnr,zhu:iccv:2017:jbo,du:pr:2018:grad}, through carefully modeling the physical characteristics of rain streaks. Benefited from large-scale training data, recent deep-learning-based derainers~\cite{fu:tip:2016:clearing,fu:cvpe:2017:ddn,yang:cvpr:2017:j,zhang:cvpr:2018:did,li:eccv:2018:rsecan,fan:acmmm:2018:rgffn,li:acmmm:2018:nled} achieve further promising performances. Nonetheless, the single-image rain removal problem remains open in two ways, as discussed below.

\begin{figure}[t]
\centering
\centering
\subfigure[Rain image]{
\includegraphics[width=0.28\linewidth]{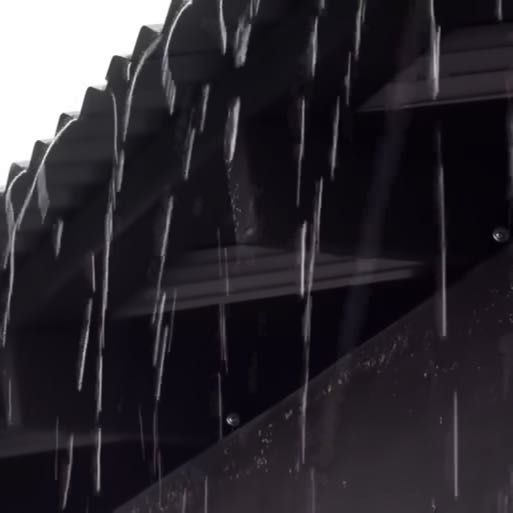}}
\centering
\vspace{-0.1in}
\subfigure[Clean image]{
\includegraphics[width=0.28\linewidth]{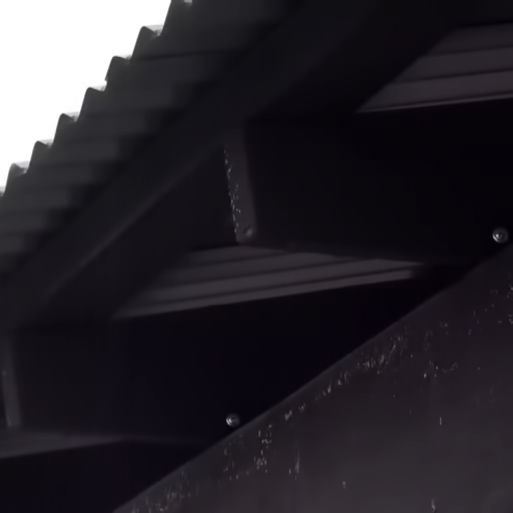}}
\centering
\subfigure[SPANet]{
\includegraphics[width=0.28\linewidth]{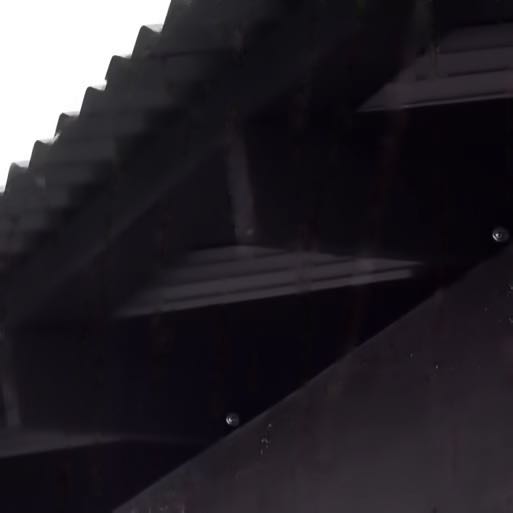}}
\subfigure[DDN~\cite{fu:cvpe:2017:ddn}]{
\includegraphics[width=0.28\linewidth]{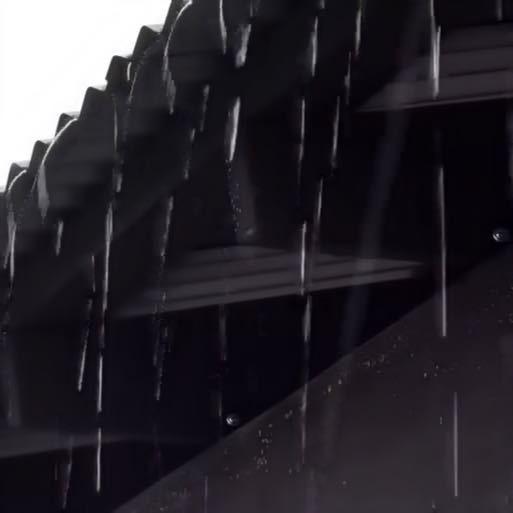}}
\centering
\subfigure[DID-MDN~\cite{zhang:cvpr:2018:did}]{
\includegraphics[width=0.28\linewidth]{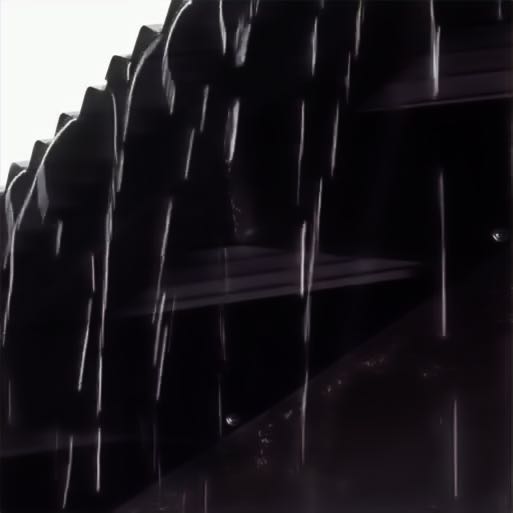}}
\centering
\subfigure[RESCAN~\cite{li:eccv:2018:rsecan}]{
\includegraphics[width=0.28\linewidth]{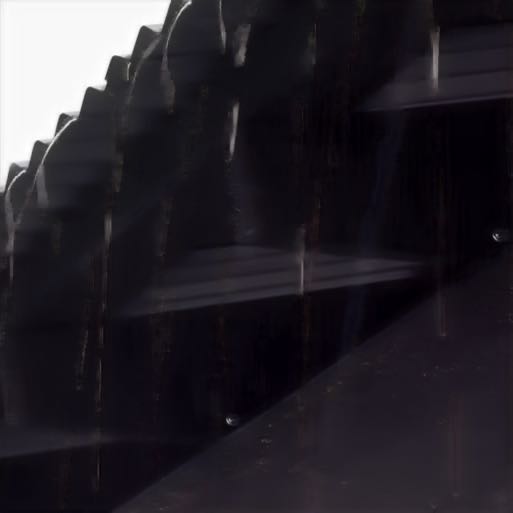}}
\centering
\caption{We address the single-image rain removal problem in two ways. First, we generate a high-quality rain/clean image pair ((a) and (b)) from each sequence of real rain images, to form a dataset. Second, we propose a novel SPANet to take full advantage of the proposed dataset. (c) to (f) compare the visual results from SPANet and from state-of-the-art derainers.}
\vspace{-5mm}
\label{fig:first}
\end{figure}

{\bf Lack of real training data.} As real rain/clean image pairs are unavailable, existing derainers typically rely on synthesized datasets to train their models. They usually start with a clean image and add synthetic rain on it to form a rain/clean image pair.
Although some works have been done to study the physical characteristics of rain, \eg, rain direction~\cite{yang:cvpr:2017:j} and rain density~\cite{zhang:cvpr:2018:did}, their datasets still lack the ability to model a large range of real world rain streaks. For example, it is often very difficult to classify the rain density into one of the three levels (\ie, light, medium and heavy) as in~\cite{zhang:cvpr:2018:did}, and any misclassification would certainly affect the deraining performance.
To simulate global rain effects, some methods adopt the nonlinear ``screen blend mode" from Adobe Photoshop, or additionally superimpose haze on the synthesized rain images. However, these global settings can only be used in certain types of rain, or the background may be darkened, with the details lost.

{\bf Lack of a real benchmark.} Currently, researchers mainly rely on qualitatively evaluating the deraining performance on real rain images through visual comparisons. Fan~\etal~\cite{fan:acmmm:2018:rgffn} also use an object detection task to help evaluate the deraining performance.
Nevertheless, a high-quality real deraining benchmark is still much needed for quantitative evaluation of deraining methods.

In this paper, we address the single-image rain removal problem in two ways, as summarized in Figure~\ref{fig:first}.
First, we address the lack of real training/evaluation datasets based on two observations:
(1) as random rain drops fall in high velocities, they unlikely cover the same pixel all the time~\cite{garg:ijcv:2007:vision,zhang:icme:2006:temporal}, and (2) the intensity of a pixel covered by rain fluctuates above the true background radiance across a sequence of images.
These two observations imply that we can generate one clean image from a sequence of rain images, where individual pixels of the clean image may be coming from different images of the sequence.
Hence, we propose a semi-automatic method that incorporates rain temporal properties as well as human supervision to construct a large-scale real rain dataset. We show that it can significantly improve the performance of state-of-the-art derainers on real world rain images. 

Second, we observe that real rain streaks can exhibit highly diverse appearance properties (\eg, rain shape and direction) within a single image, which challenges existing derainers as they lack the ability to identify real rain streaks accurately.
To address this limitation, we exploit a spatial attentive network (SPANet), which first leverages horizontal/vertical neighborhood information to model the physical properties of rain streaks, and then remove them by further considering the non-local contextual information.
In this way, the discriminative features for rain streak removal can be learned in a two-stage local-to-global manner.
Extensive evaluations show that the proposed network performs favorably against the state-of-the-art derainers.

To summarize, this work has the following contributions:
\begin{enumerate}
	\item We present a semi-automatic method that incorporates temporal properties of rain streaks and human supervision to generate a high quality clean image from a sequence of real rain images.

	\item We construct a large-scale dataset of $\sim$$29.5K$ high-resolution rain/clean image pairs, which covers a wide range of natural rain scenes. We show that it can significantly improve the performance of state-of-the-art derainers on real rain images.

  \item We design a novel SPANet to effectively learn discriminative deraining features in a local-to-global attentive manner. SPANet achieves superior performance over state-of-the-art derainers.

\end{enumerate}

\section{Related works}

{\bf Single-image rain removal.} This problem is extremely challenging due to the ill-posed deraining formulation as:
\begin{equation}\label{rmod}
B = O - R,
\end{equation}
where $O$, $R$ and $B$ are the input rain image, the rain streak image, and the output derained image, respectively.

Kang~\etal~\cite{kang:tip:2012:imgdecomp} propose to first decompose the rain image into high-/low-frequency layers and remove rain streaks in the high frequency layer via dictionary learning.
Kim~\etal~\cite{kim:tip:2015:videoderain} propose to use non-local mean filters to filter out rain streaks.
Luo~\etal~\cite{luo:iccv:2015:dsc} propose a sparse coding based method to separate rain streaks from the background.
Li~\etal~\cite{li:cvpr:2016:lp} propose to use Gaussian mixture models to model rain streaks and background separately for rain removal.
Chang~\etal~\cite{chang:iccv:2017:lpnr} propose to first affine transform the rain image into a space where rain streaks have vertical appearances and then utilize the low-rank property to remove rain streaks.
Zhu~\etal~\cite{zhu:iccv:2017:jbo} exploit rain streak directions to first determine the rain-dominant regions, which are used to guide the process of separating rain streaks from background details based on rain-dominant patch statistics.

In~\cite{fu:cvpe:2017:ddn,fu:tip:2016:clearing}, deep learning is applied to single image deraining and achieves a significant performance boost. They model rain streaks as ``residuals" between the input/output of the networks in an end-to-end manner.
Yang~\etal~\cite{yang:cvpr:2017:j} propose to decompose the rain layer into a series of sub-layers representing rain streaks of different directions and shapes, and jointly detect and remove rain streaks using a recurrent network.
In~\cite{zhang:arxiv:2017:gan}, Zhang~\etal propose to remove rain streaks and recover the background via the Conditional GAN.
Recently, Zhang and Patel~\cite{zhang:cvpr:2018:did} propose to classify rain density to guide the rain removal step.
Li~\etal~\cite{li:eccv:2018:rsecan} propose a recurrent network with a squeeze-and-excitation block~\cite{hu:cvpr:2018:senet} to remove rain streaks in multiple stages.
However, the performances of CNN-based derainers on real rain images are largely limited by being trained only on synthetic datasets. These derainers also lack the ability to attend to rain spatial distributions. In this paper, we propose to leverage real training data as well as a spatial attentive mechanism to address the single image deraining problem.

{\bf Multi-image rain removal.} Unlike single-image deraining, rich temporal information can be derived from a sequence of images to provide additional constraints for rain removal. Pioneering works~\cite{garg:cvpr:2004:detection,garg:ijcv:2007:vision} propose to apply photometric properties to detect rain streaks and estimate the corresponding background intensities by averaging the irradiance of temporal or spatial neighboring pixels. Subsequently, more intrinsic properties of rain streaks, such as chromatic property, are explored by~\cite{zhang:icme:2006:temporal,liu:cis:2009:pixel,santhaseelan:ijcv:2015:utilizing}. Recent works~\cite{bossu:ijcv:2011:raindetection,chen:iccv:2013:generalized,chen:tip:2014:rpra,kim:tip:2015:videoderain,jiang:cvpr:2017:dip,ren:cvpr:2017:md,wei:iccv:2017:ds,li:cvpr:2018:mcsc,liu:cvpr:2018:j4r} focus on removing the rain streaks from the background with moving objects. Chen~\etal~\cite{chen:cvpr:2018:videoderaincnn} further propose a spatial-temporal content alignment algorithm to handle fast camera motion and dynamic scene contents, and a CNN to reconstruct high frequency background details.

However, these methods cannot be applied for our purpose of generating high-quality rain-free images. This is because if their assumptions (\eg, low-rank~\cite{chen:iccv:2013:generalized,wei:iccv:2017:ds,li:cvpr:2018:mcsc}) are violated, over-/under-deraining can happen to the entire sequence and further bury the true background radiance, \ie, the clean background pixels may not exist in this sequence. Hence, in this paper, we propose to use the original sequence of rain images to generate a clean image, and rely on human judgements on the qualities of generated rain-free images.

{\bf Generating the ground truth from real noisy images.} One typical strategy~\cite{anaya:arxiv:2014:RENOIR,plotz:cvpr:2017:noisebench} to obtain a noise/noise-free image pair is to photograph the scene with a high ISO value and a short exposure time for the noise image, and a low ISO value and a long exposure time for the noise-free image. However, this strategy cannot be used here to capture rain-free images. As rain drops fall at a high speed, increasing the exposure time will enlarge the rain streaks, not removing them.
Another approach to obtain a ground truth noise-free image is multi-frame fusion~\cite{zhu:cvpr:2016:denoise,nam:cvpr:2016:hadenoise,abdelhamed:cvpr:2018:dndata},
which performs weighted averaging of a pre-aligned sequence of images taken from a static scene with a fixed camera setting.
However, as rain streaks have brighter appearances and larger shapes than random noise, this approach is not able to accurately remove rain from the rain pixels.
In contrast, we propose to refine the rain pixels based on the observation that the intensity values of the pixels covered by rain fluctuate above their true background intensities.

\section{Real Rain Image Dataset}\label{sec:dataset}

\begin{figure}[h]
\begin{minipage}[t]{\linewidth}
\includegraphics[width=\textwidth]{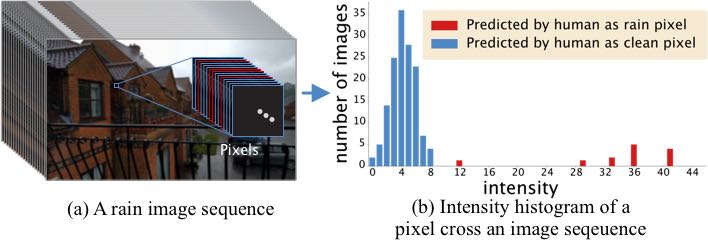}
\end{minipage}
\caption{We trace the intensity of one pixel across an image sequence in (a). We ask a user to identify if this pixel in each frame is covered by rain (in red) or not (in blue). The intensity distribution of this pixel over all frames is show in (b). It shows that the intensity of the pixel tends to fluctuate in a smaller range if it is not covered by rain, as compared with that covered by rain.}
\label{fig:mode}
\end{figure}

We first conduct an experiment on how to select a suitable background value $o_b$ from a collection of pixel values $O_l = \{ o_{1l},...,o_{Nl}\}$ at spatial position $l$ from a sequence of $N$ rain images. We capture a video of a rain scene over a static background, as shown in Figure~\ref{fig:mode}, and then ask a person to indicate (or predict) when a particular pixel is covered by rain and when it is not, across the $N$ frames. We have observed two phenomena.
First, rain streaks do not always cover the same pixel (the temporal property of video deraining~\cite{zhang:icme:2006:temporal}).
Second, humans typically predict if a pixel is covered by rain or not based on the pixel intensity. If the intensity of the pixel is lower at a certain frame compared with the other frames, humans would predict that it is not covered by rain. This is because rain streaks tend to brighten the background. These two observations imply that, given a sequence of $N$ consecutive rain images, we can approximate the true background radiance $B_{l}$ at pixel $l$ based on these human predicted rain-free pixel values (i.e., the blue region of the histogram in Figure~\ref{fig:mode}(b)). If we assume that the ambient light is constant during this time span, we can then use the value that appears most frequently (i.e., mode in statistics) to approximate the background radiance.

\begin{figure*}[ht]
\begin{minipage}[t]{0.67\textwidth}
\centering
\includegraphics[width=0.95\textwidth]{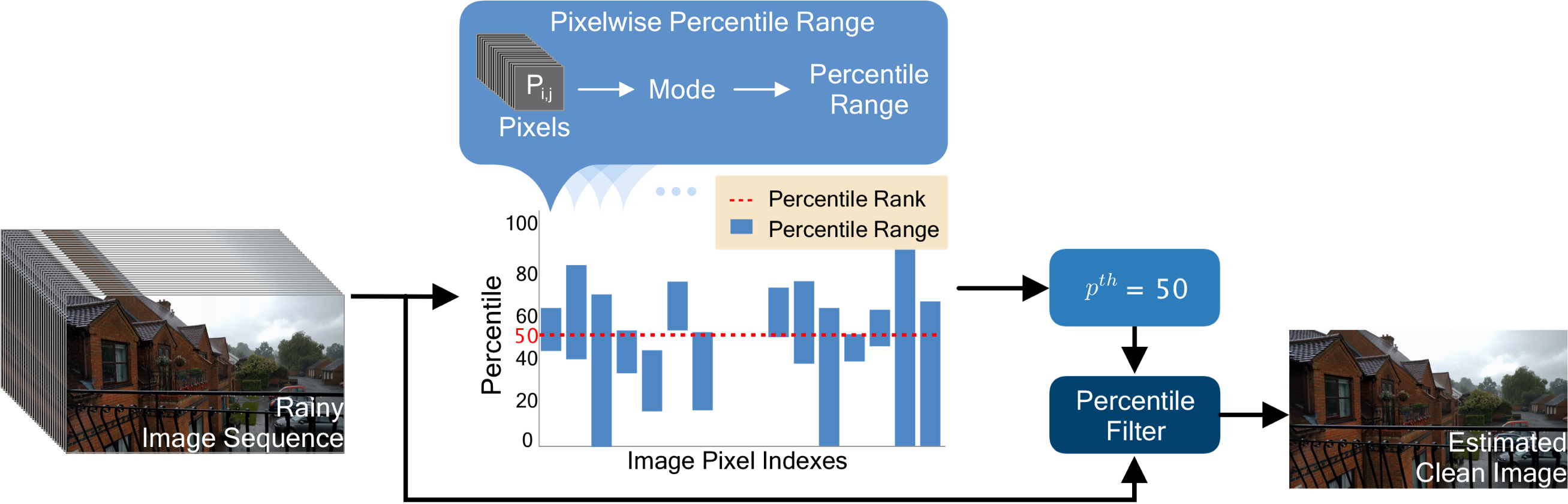}
\centerline{\footnotesize (a) Pipeline of Background Approximations}
\end{minipage}
\begin{minipage}[t]{0.32\textwidth}
\centering
\includegraphics[width=\textwidth]{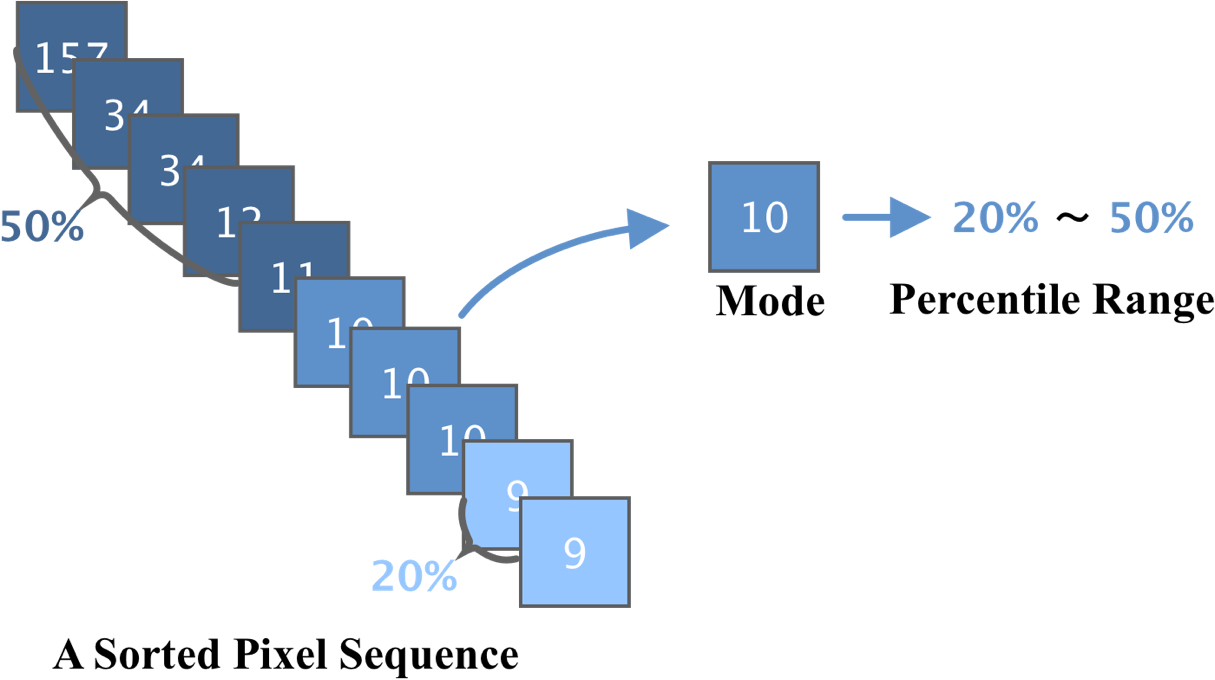}
\centerline{\footnotesize (b) Computing Percentile Range for Mode Value}
\end{minipage}
\vspace{0.05in}
\caption{Overview of our clean image generation pipeline (a). Given a sequence of rain images, we compute the mode for each pixel based on its intensity changes over time, and the percentile range of its mode. We then consider the global spatial smoothness by finding a percentile rank that can cross most of the percentile ranges (b).}
	\label{fig:sapf}
	\vspace{-3mm}
\end{figure*}

{\bf Background approximation.} Referring to Figure~\ref{fig:sapf}, given a set of pixel values $O_l$ at position $l$ from a sequence of $N$ rain images, we first compute the mode of $O_l$ as:
\begin{equation}
\phi_l = \Phi(O_l), \label{eq:mode}
\end{equation}
where $\Phi$ is the mode operation. However, since Eq.~\ref{eq:mode} does not consider the neighborhood information when computing $\phi_l$, the resulting images tend to be noisy in dense rain streaks. So, we identify the percentile range ($R_l^{min}$, $R_l^{max}$) of the computed $\phi_l$ in $O_l$ based on their intensity values as:
\begin{eqnarray}
\nonumber
R_l^{min} = \frac{100\%}{N}\sum_{i=1}^N \{1|o_{il} < \phi_l\},\\
R_l^{max} = \frac{100\%}{N}\sum_{i=1}^N \{1|o_{il} \leq \phi_l\}.
\end{eqnarray}

\begin{figure*}[t]
	\centering
		\subfigure[Input]{
			\includegraphics[width = 0.125\linewidth]{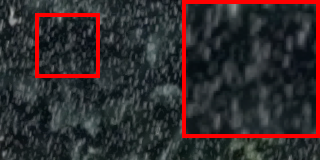}
		}
		\subfigure[Jiang~\cite{jiang:cvpr:2017:dip}]{
			\includegraphics[width = 0.125\linewidth]{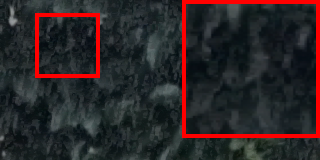}
		}
		\subfigure[Wei~\cite{wei:iccv:2017:ds}]{
			\includegraphics[width = 0.125\linewidth]{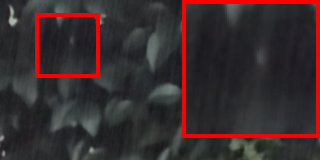}
		}
		\subfigure[Li~\cite{li:cvpr:2018:mcsc}]{
			\includegraphics[width = 0.125\linewidth]{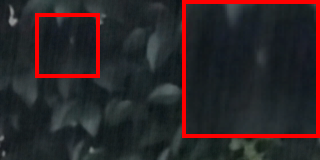}
		}
		\subfigure[Mode filter]{
			\includegraphics[width = 0.125\linewidth]{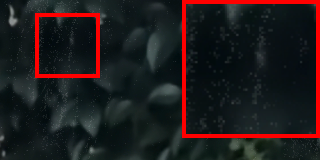}
		}
		\subfigure[\textbf{Ours}]{
			\includegraphics[width = 0.125\linewidth]{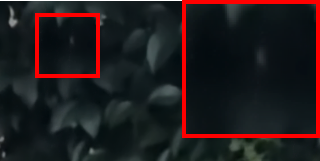}
		}
		\subfigure[Ground Truth]{
			\includegraphics[width = 0.125\linewidth]{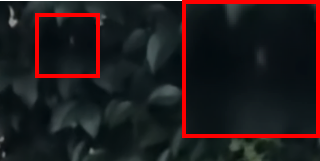}
		}
		\caption{A deraining example using a synthetic rain video of 100 frames. We show the best result of each method here. Refer to the supplementary for more results. }
		\label{fig:example1}
		\vspace{-3mm}
\end{figure*}

Figure~\ref{fig:sapf}(c) shows an example. Instead of using polygonal lines to connect the mode values $\phi_l$ at all spatial positions,
we can determine a suitable percentile $\hat{p}$ so that it crosses the highest number of percentile ranges (the red dash line in Figure~\ref{fig:sapf}(c)).
In this way, the estimated background image is globally smoothed by computing $\hat{p}$ as:
\begin{equation}
\begin{aligned}
\hat{p} = \arg\max_{p}(\{\sum_{l=0}^{M-1} \{1 |R_l^{min} < p<  R_l^{max}\}\}_{p=0}^{100}),
\end{aligned}
\end{equation}
where $M$ is the number of pixels in a frame. Figure~\ref{fig:example1}(e) shows an example that using the mode leads to noisy result, while our method in Figure~\ref{fig:example1}(f) produces a cleaner image.

{\bf Selection of $N$ for different rain scenes.} Recall that we aim to generate one clean image from a sequence of $N$ rain images. Our method assumes that for each pixel of the output clean image, we are able to find some input frames where the pixel is not covered by rain. To satisfy this assumption, we need to adjust $N$ according to the amount of rain as follows.
First, we empirically set $N$ to be $\{20, 100, 200\}$ depending on whether the rain is $\{sparse, normal, dense\}$, respectively, and generate an output image using our method.
Second, we ask users to evaluate the image as humans are sensitive to rain streaks as well as other artifacts such as noise.
If the image fails in the user evaluation, we adjust $N$ by adding $\{10, 20, 50\}$ frames for $\{sparse, normal, dense\}$ rain streaks and then ask the users to evaluate the new output image again.
We find that while $20$ and $100$ frames are usually large enough to obtain a clean image for $sparse$ and $normal$ rain streaks, $N$ may go from $200$ to $300$ frames for $dense$ rain streaks. We deliberately start with smaller numbers of frames because we find that the more frames that we use, the higher chance that the video may contain noise, blur and shaking.

{\bf Discussion.} An intuitive alternative to obtaining a rain-free image is to use a state-of-the-art video deraining method to first generate a sequence of derained results from the input rain sequence, and then average them or select the best result from them to produce a single final rain-free image. Unfortunately, there is no guarantee that rain streaks can be completely removed by the video deraining method, as shown in Figure~\ref{fig:example1}(b)-(d). 
On the contrary, we rely on human judgements to generate high-quality rain-free images.
We show a comparison between our method and three state-of-the-art video deraining methods~\cite{jiang:cvpr:2017:dip,wei:iccv:2017:ds,li:cvpr:2018:mcsc} in Table~\ref{table:video} on 10 synthesized rain videos (10 black-background rain videos bought from \cite{rainstreaks} are imposed on 10 different background images), which clearly demonstrates the effectiveness of our method.

\begin{table}[t]
\centering
\resizebox{\linewidth}{!}{%
\begin{tabular}{@{}cccccc@{}}
\toprule
Methods & Input & Jiang~\etal~\cite{jiang:cvpr:2017:dip} & Wei~\etal~\cite{wei:iccv:2017:ds} & Li~\etal~\cite{li:cvpr:2018:mcsc} & Ours \\ \midrule
PSNR & 25.40  & 32.79 (29.82)   & 27.30 (25.71)   & 32.59 (30.59)  & \textbf{51.40} \\
SSIM & 0.7228 & 0.8827 (0.8566) & 0.9043 (0.8911) &0.9458 (0.9387) & \textbf{0.9907} \\
\bottomrule
\end{tabular}%
}
\caption{Comparison with the state-of-the-art video deraining methods. In each method, we select the frame of highest PSNR for comparison. The average PSNR/SSIM are in brackets.}
\label{table:video}
\vspace{-4mm}
\end{table}

{\bf Dataset description.} We construct a large-scale dataset using $170$ real rain videos, of which $84$ scenes are captured by us using an iPhone X or iPhone 6SP
and $86$ scenes are collected from StoryBlocks or YouTube.
These videos cover common urban scenes (\eg, buildings, avenues), suburb scenes (\eg, streets, parks), and some outdoor fields (\eg, forests). When capturing rain scenes, we also control the exposure durations as well as the ISO parameter to cover different lengths of rain streaks and illumination conditions.
Using the aforementioned method, we generate $29,500$ high-quality rain/clean image pairs, which are split into $28,500$ for training and $1,000$ for testing. Our experiments show that this dataset helps improve the performance of state-of-the-art derainers.

\begin{figure*}[htbp]
\begin{minipage}[t]{\textwidth}
\centering
\includegraphics[height=28pt]{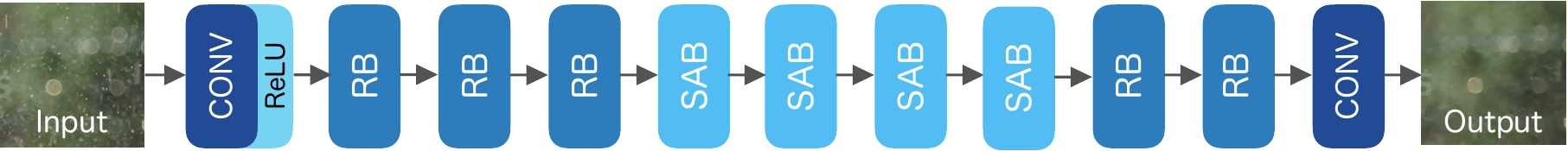}
\centerline{\footnotesize (a) Spatial Attentive Network \textbf{(SPANet)}}
\end{minipage}
\space{}
\begin{minipage}[t]{0.4\textwidth}
\centering
\includegraphics[width=0.75\textwidth]{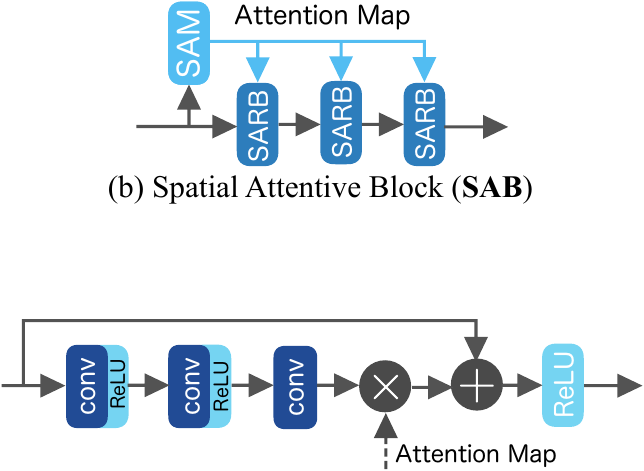}
\centerline{\footnotesize (c) Spatial Attentive Residual Block \textbf{(SARB)}}
\end{minipage}
\begin{minipage}[t]{0.6\textwidth}
\centering
\includegraphics[width=0.89\textwidth]{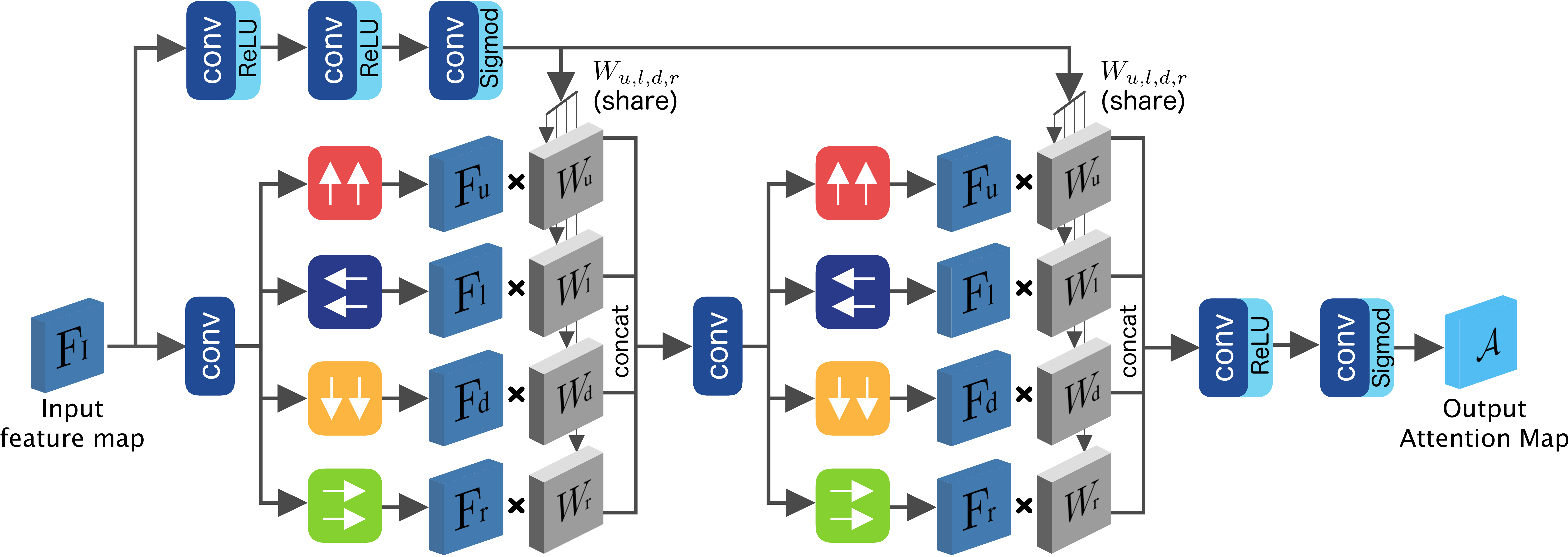}	
\centerline{\footnotesize (d) Spatial Attentive Module \textbf{(SAM)}}
\end{minipage}
\space{\ }
\caption{The architecture of the proposed \textbf{SPANet} (a). It adopts three standard residual blocks ({\bf RB}s)~\cite{he:cvpr:2016:resnet} to extract features, four spatial attentive blocks ({\bf SAB}s) to identify rain streaks progressively in four stages, and two residual blocks to reconstruct a clean background. A {\bf SAB} (b) contains three spatial attentive residual blocks ({\bf SARB}s) (c) and one spatial attentive module ({\bf SAM}) (d). Dilation convolutions~\cite{yu:iclr:2015:dilated} are used in {\bf RB}s and {\bf SARB}s.}
\label{fig:network}
\vspace{-3mm}
\end{figure*}

\section{Proposed Model}\label{sec:spanet}

As real rain streaks may have highly diverse appearances across the image, we propose the SPANet to detect and remove rain streaks in a local-to-global manner, as shown in Figure~\ref{fig:network}(a). It is a fully convolutional network that takes one rain image as input and outputs a derained image.

\subsection{Spatial Attentive Block}

{\bf Review on IRNN architecture.} Recurrent neural networks with ReLU and identity matrix initialization (IRNN) for natural language processing~\cite{le:arxiv:2015:simple} have been shown to be easy to train, good at modeling long-range dependencies as well as efficient. When applied to computer vision problems, their key advantage is that information can be efficiently propagated across the entire image to accumulate long range varying contextual information, by stacking at least two RNN layers.
In~\cite{bell:cvpr:2016:ion}, a two-round four-directional IRNN architecture is used to exploit contextual information to improve small object detection. While the first round IRNN aims to produce the feature maps that summarize the neighboring contexts for each position of the input image, the second round IRNN further gathers non-local contextual information for producing global aware feature maps.
Recently, Hu~\etal~\cite{hu:cvpr:2018:dsc} also exploit this two-round four-directional IRNN architecture to detect shadow regions based on the observation that directions play an important role in finding strong cues between shadow/non-shadow regions. They design a direction-aware attention mechanism to generate more discriminative contextual features.

\begin{figure}[t]
\centering
\includegraphics[width=0.9\linewidth]{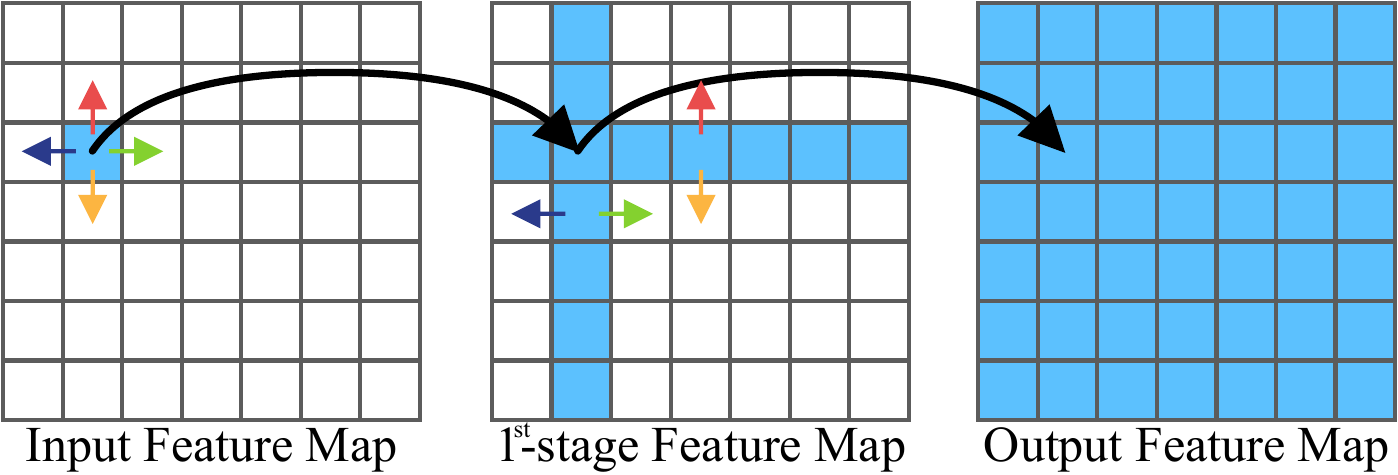}
\caption{Illustration of how the two-round four-directional IRNN architecture accumulates global contextual information in two stages. In the first stage, for each position at the input feature map, four-directional (up, left, down, right) recurrent convolutional operations are performed to collect horizontal and vertical neighborhood information. In the second stage, by repeating the previous operations, the contextual information from the entire input feature map are obtained.}
\label{fig:GA}
\vspace{-4mm}
\end{figure}

We summarize the four-directional IRNN operation for computing feature $h _ { i , j }$ at location $(i,j)$ as:
\begin{equation}\label{eq:trans}
  h _ { i , j }  \leftarrow \max \left( \alpha_{dir} \ h _ { i , j - 1 }  + h _ { i , j }  , 0 \right) ,
\end{equation}
where $\alpha_{dir}$ denotes the weight parameter in the recurrent convolution layer for each direction. Figure~\ref{fig:GA} illustrates how a two-round four-directional IRNN architecture accumulates global contextual information.
Here, we extend the two-round four-directional IRNN model to the single-image rain removal problem, for the purpose of handling the significant appearance variations of real rain streaks.

\begin{figure}[h]
\begin{minipage}[t]{\linewidth}
\ \ \ \ \ \ \ \ \ \ \ \ \ \ \ \ \ \ \ \ \ \ \ \ \ \includegraphics[width=0.48\linewidth]{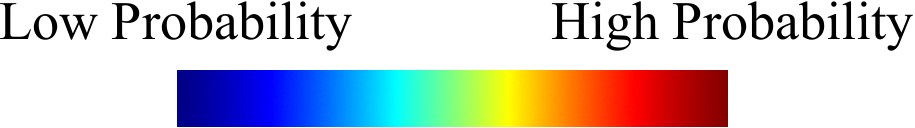}
\vspace{1mm}
\end{minipage}
\centering
\begin{minipage}[t]{0.32\linewidth}
\centering
\includegraphics[width=0.6\textwidth]{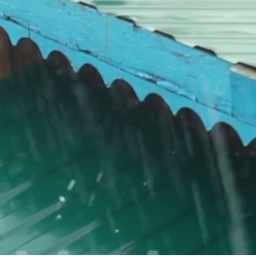}
\end{minipage}
\centering
\begin{minipage}[t]{0.32\linewidth}
\centering
\includegraphics[width=0.6\textwidth]{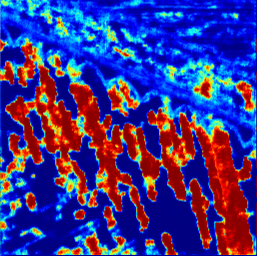}
\end{minipage}
\begin{minipage}[t]{0.32\linewidth}
\centering
\includegraphics[width=0.6\textwidth]{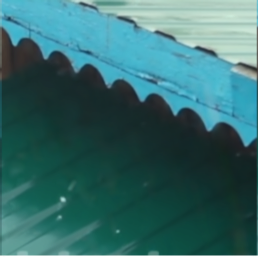}
\vspace{-1mm}
\end{minipage}
\centering
\begin{minipage}[t]{0.32\linewidth}
\centering
\centerline{\footnotesize (a) Rain Image}
\end{minipage}
\centering
\begin{minipage}[t]{0.32\linewidth}
\centering
\centerline{\footnotesize (b) Attention Map}
\end{minipage}
\begin{minipage}[t]{0.32\linewidth}
\centering
\centerline{\footnotesize (c) SPANet Result}
\vspace{0.1mm}
\end{minipage}

\vspace{0mm}
\caption{Visualization of the attention map. (a) shows one real rain image. (b) shows the corresponding attention map produced by SAM. Red color indicates pixels that are highly likely covered by rain. (c) shows the corresponding derained result by the proposed SPANet. This demonstrates the effectiveness of SAM in handling significant appearance variations of rain streaks. Refer to the supplementary for more results.}
\label{fig:att}
\vspace{-3mm}
\end{figure}

{\bf Spatial attentive module (SAM).} We build SAM based on the aforementioned two-round four-directional IRNN architecture. We use the IRNN model to project the rain streaks to the four main directions. Another branch is added to capture the spatial contextual information in order to selectively highlight the projected rain features, as shown in Figure~\ref{fig:network}(d).
Unlike~\cite{hu:cvpr:2018:dsc} that implicitly learns direction-aware features in the embedding space, we further use additional convolutions and sigmoid activations to explicitly generate the attention map through explicit supervision. The attention map indicates rain spatial distributions and is used to guide the following deraining process. 
Figure~\ref{fig:att} shows the input rain images in (a) and our SPANet derained results in (c). We also visualize the attention maps produced by SAM in (b). We can see that SAM can effectively identify the regions affected by rain streaks, even though the rain streaks exhibit significant appearance variations (\ie, smooth and blurry in the first scene and sharp in the second scene).

{\bf Removal-via-detection.}  As shown in Figure~\ref{fig:network}(a), given an input rain image, three standard residual blocks (RBs)~\cite{he:cvpr:2016:resnet} are first used to extract features. We feed these features into a spatial attentive block (SAB) (Figure~\ref{fig:network}(b)), which uses a SAM to generate an attention map to guide three subsequent spatial attentive residual blocks (SARBs) (Figure~\ref{fig:network}(c)) to remove rain streaks via the learned negative residuals. The SAB is repeated four times. (Note that the weights of the SAM in the four SABs are shared.) Finally, the resulting feature maps are fed to two standard residual blocks to reconstruct the final clean background image.

\subsection{Training Details}~\label{sec:train}

{\bf Loss function.} We adopt the following loss function to train SPANet:
\begin{equation}
\mathcal { L }_{total} = \mathcal { L }_{1}  + \mathcal { L }_{SSIM} +\mathcal { L }_{Att}.
\end{equation}
We use the standard $\mathcal{L}_{1}$ loss to measure the per-pixel reconstruction accuracy. $\mathcal{L}_{ssim}$~\cite{wang:tip:2004:ssim} is used to constrain the structural similarities, and is defined as: $1 - SSIM(\mathcal{P},\mathcal{C})$,
where $\mathcal{P}$ is the predicted result and $\mathcal{C}$ is the clean image. 
We further apply the attention loss $\mathcal{L}_{att}$ as:
\begin{equation}
\mathcal { L }_{att} = \left\| \mathcal{A} - \mathcal{M}\right\| _{2} ^{2},
\end{equation}
where $\mathcal{A}$ is the attention map from the first SAM in the network and $\mathcal{M}$ is the binary map of the rain streaks, which is computed by thresholding the difference between the rain image and clean image. In this binary map, a $1$ indicates that the pixel is~covered by rain and $0$ otherwise.

\begin{table*}[bp]
\centering
\resizebox{\textwidth}{!}{%
\begin{tabular}{@{}ccccc|cccccc@{}}
\toprule
Methods & \begin{tabular}[c]{@{}c@{}}Rain \\ Images\end{tabular} & \begin{tabular}[c]{@{}c@{}}DSC \cite{luo:iccv:2015:dsc}\\ (ICCV'15) \end{tabular} & \begin{tabular}[c]{@{}c@{}}LP \cite{li:cvpr:2016:lp}\\ (CVPR'16) \end{tabular} & \begin{tabular}[c]{@{}c@{}}SILS \cite{gu:iccv:2017:sils}\\ (ICCV'17) \end{tabular}& \begin{tabular}[c]{@{}c@{}}Clear \cite{fu:tip:2016:clearing}\\ (TIP'17) \end{tabular}  & \begin{tabular}[c]{@{}c@{}}DDN \cite{fu:cvpe:2017:ddn} \\ (CVPR'17) \end{tabular}  & \begin{tabular}[c]{@{}c@{}}JORDER \cite{yang:cvpr:2017:j}\\ (CVPR'17) \end{tabular} &  \begin{tabular}[c]{@{}c@{}}DID-MDN \cite{zhang:cvpr:2018:did}\\ (CVPR'18) \end{tabular} &  \begin{tabular}[c]{@{}c@{}}RESCAN \cite{li:eccv:2018:rsecan}\\(ECCV18) \end{tabular}  &  \begin{tabular}[c]{@{}c@{}}\textbf{Our}\\ \textbf{SPANet}\end{tabular} \\ \midrule
PSNR & 32.64 & 32.33 & 32.99 & 33.40 &  31.31 & 33.28 (\red{34.88})  & 32.16 (\red{35.72}) & 24.91 (\red{28.96})& 30.36 (\red{35.19})  & {\bf 38.06} \\ \midrule
SSIM & 0.9315 & 0.9335 & 0.9475 & 0.9528  & 0.9304 & 0.9414 (\red{0.9727})  & 0.9327 (\red{0.9776})&  0.8895(\red{0.9457}) & 0.9553 (\red{0.9784}) & {\bf 0.9867} \\ \bottomrule
\end{tabular}
}
\caption{Quantitative results for benchmarking the proposed SPANet and the state-of-the-art derainers on the proposed test set. The original codes of all these derainers are used for evaluation. We have also trained CNN-based state-of-the-art methods~\cite{fu:cvpe:2017:ddn,yang:cvpr:2017:j,zhang:cvpr:2018:did,li:eccv:2018:rsecan} on our dataset, and results are marked in \red{red}. The best performance is marked in {\bf bold}. Note that due to the lack of density labels for the rain images in our dataset, we only fine-tune the pre-trained model of DID-MDN~\cite{zhang:cvpr:2018:did} without the re-training label classification network.}
\label{table:benchmark}

\end{table*}

{\bf Implementation details.} SPANet is implemented using the PyTorch~\cite{pytorch} framework on a PC with a E5-2640 v4 2.4GHz CPU and 4 NVIDIA Titan V GPUs. For loss optimization, we adopt the Adam optimizer~\cite{kingma:iclr:2014:adam} with a batch size of 16.~We adopt scaling and cropping to augment the diversity of rain streaks.
The learning rate is initialized at 0.005 and divided by 10 after 30K iterations. We train the network for 40K iterations.

\section{Experiments}
In this section, We first evaluate the effectiveness of the proposed dataset on existing CNN-based single-image derainers, and then compare the proposed SPANet to the state-of-the-art single-image deraining methods. Finally, we provide internal analysis to study the contributions of individual components of SPANet. Refer to the supplementary for more results.

\begin{figure*}[t]
\centering
\begin{minipage}[t]{0.85\textwidth}
\centering
\includegraphics[width=\textwidth]{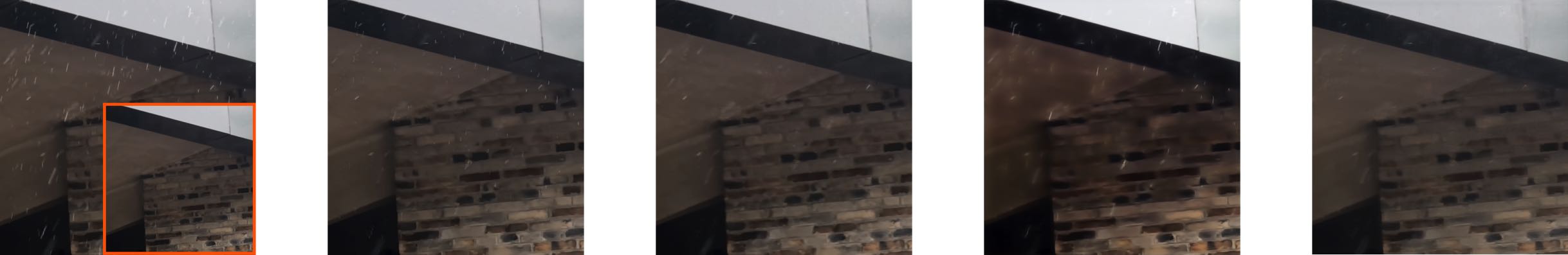}
\end{minipage}

\begin{minipage}[t]{0.17\textwidth}
\centerline{\footnotesize (a) Rain / Clean Image}
\centerline{\footnotesize 33.53 / 0.9372 }
\end{minipage}
\centering
\begin{minipage}[t]{0.17\linewidth}
\centerline{\footnotesize (b) DDN~\cite{fu:cvpe:2017:ddn}}
\centerline{\footnotesize 37.27 / 0.9631}
\end{minipage}
\begin{minipage}[t]{0.17\linewidth}
\centerline{\footnotesize (c) JORDER~\cite{yang:cvpr:2017:j}}
\centerline{\footnotesize 36.67 / 0.9657 }
\end{minipage}
\begin{minipage}[t]{0.17\linewidth}
\centerline{\footnotesize (d) DID-MDN~\cite{zhang:cvpr:2018:did}}
\centerline{\footnotesize 22.86 / 0.8721 }
\end{minipage}
\begin{minipage}[t]{0.17\linewidth}
\centerline{\footnotesize (e) RESCAN~\cite{li:eccv:2018:rsecan}}
\centerline{\footnotesize 35.80 / 0.9538}
\end{minipage}

\begin{minipage}[t]{\textwidth}
\centerline{  }
\end{minipage}

\begin{minipage}[t]{0.85\textwidth}
\includegraphics[width=\textwidth]{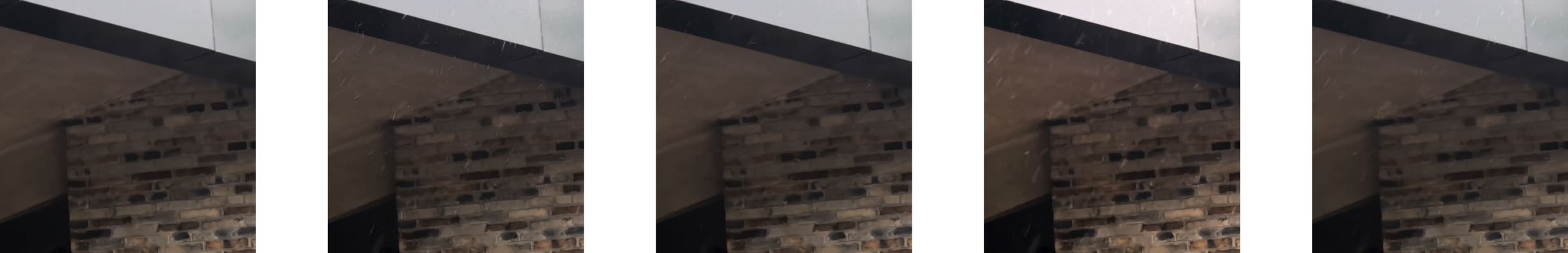}
\end{minipage}

\begin{minipage}[t]{0.17\linewidth}
\centerline{\footnotesize \textbf{(f)} { \textbf{Our SPANet}}}
\centerline{\footnotesize \textbf{43.49} / \textbf{0.9938}}
\end{minipage}
\centering
\begin{minipage}[t]{0.17\textwidth}
\centerline{\footnotesize (g) {\color[HTML]{FF0000}DDN}~\cite{fu:cvpe:2017:ddn}}
\centerline{\footnotesize 38.36 / 0.9668 }
\end{minipage}
\centering
\begin{minipage}[t]{0.17\textwidth}
\centerline{\footnotesize (h) {\color[HTML]{FF0000}JORDER}~\cite{yang:cvpr:2017:j}}
\centerline{\footnotesize 40.49 / 0.9834}
\end{minipage}
\centering
\begin{minipage}[t]{0.17\textwidth}
\centerline{\footnotesize (i) {\color[HTML]{FF0000}DID-MDN}~\cite{zhang:cvpr:2018:did}}
\centerline{\footnotesize 26.54 / 0.9625}
\end{minipage}
\begin{minipage}[t]{0.17\textwidth}
\centerline{\footnotesize (j) {\color[HTML]{FF0000}RESCAN}~\cite{li:eccv:2018:rsecan}}
\centerline{\footnotesize 39.29 / 0.9771}
\end{minipage}
\vspace{1.5mm}
\caption{Visual comparison of the state-of-the-art CNN-based derainers trained on the original/proposed datasets. Methods in \red{red} mean that they are retrained on the proposed dataset. PSNR/SSIM results are included for reference.}
\label{fig:dataset}
\end{figure*}

{\bf Evaluation on the proposed dataset.} The performances of existing CNN-based derainers~\cite{fu:cvpe:2017:ddn,yang:cvpr:2017:j,zhang:cvpr:2018:did,li:eccv:2018:rsecan} trained on our dataset are shown in Table~\ref{table:benchmark}. It demonstrates that our real dataset can significantly improve the performance of CNN-based methods on real images. This is mainly due to the fact that existing synthesized datasets lack the ability to represent highly varying rain streaks. One visual example is given in Figure~\ref{fig:dataset}, from which we can see that the retrained derainers can produce cleaner images with more details compared to those trained on synthetic datasets. Note that we use their original codes for evaluation and retraining.

\begin{figure}[ht]
	\centering
		\begin{minipage}[t]{0.15\linewidth}
		\includegraphics[width=\textwidth]{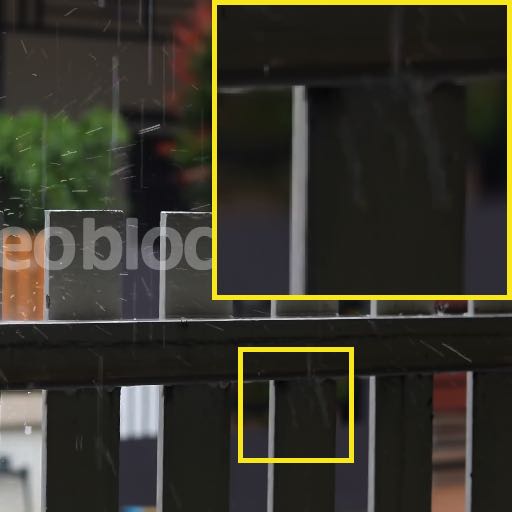}\vspace{-0.08in}
		\vspace{-0.08in}\centerline{\tiny Rain}
		\centerline{\tiny 34.2}
		\end{minipage}
		\begin{minipage}[t]{0.15\linewidth}
		\includegraphics[width=\textwidth]{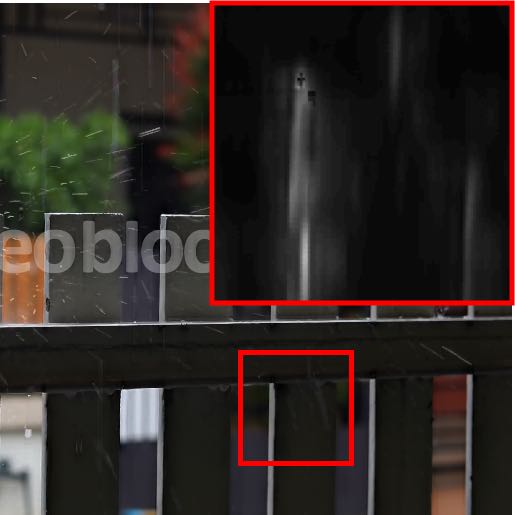}\vspace{-0.08in}
		\vspace{-0.08in} \centerline{\tiny DSC {\cite{luo:iccv:2015:dsc}}}
		\centerline{\tiny 30.9}
		\end{minipage}
		\begin{minipage}[t]{0.15\linewidth}
		\includegraphics[width=\textwidth]{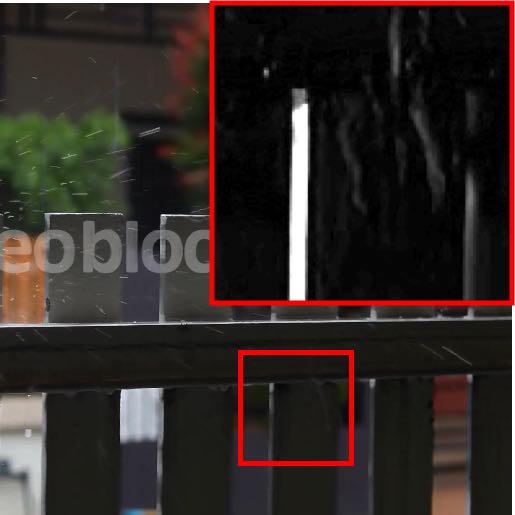}\vspace{-0.08in}
		\vspace{-0.08in} \centerline{\tiny JORDER {\cite{yang:cvpr:2017:j}}}
		\centerline{\tiny 27.2}
		\end{minipage}
		\begin{minipage}[t]{0.15\linewidth}
		\includegraphics[width=\textwidth]{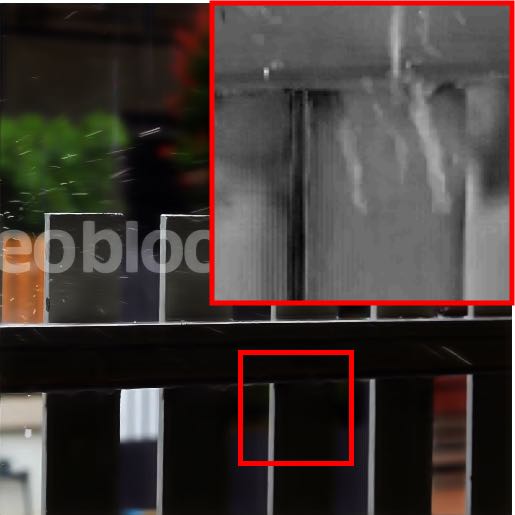}\vspace{-0.08in}
		\vspace{-0.08in} \centerline{\tiny DID-MDN {\cite{zhang:cvpr:2018:did}}}
		\centerline{\tiny 23.0}
		\end{minipage}
		\begin{minipage}[t]{0.15\linewidth}
		\includegraphics[width=\textwidth]{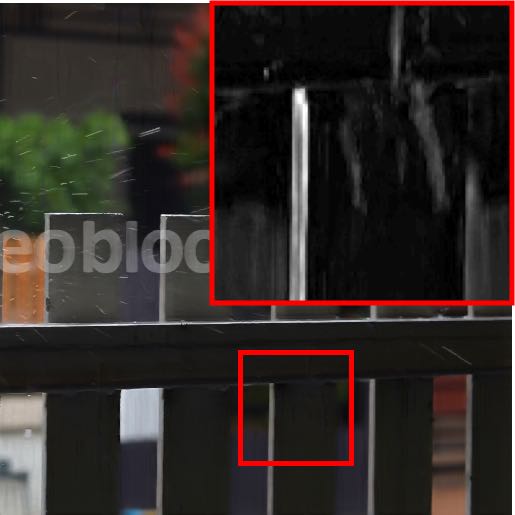}\vspace{-0.08in}
		\vspace{-0.08in}\centerline{\tiny RESCAN {\cite{li:eccv:2018:rsecan}}}
		\centerline{\tiny 31.1}
		\end{minipage}
		\begin{minipage}[t]{0.15\linewidth}
		\includegraphics[width=\textwidth]{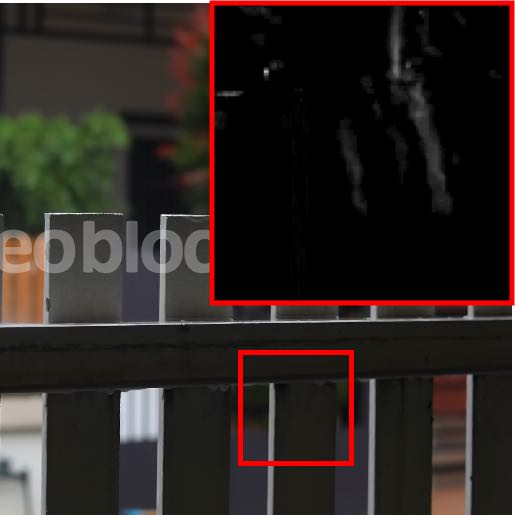}\vspace{-0.08in}
		\vspace{-0.08in}\centerline{\tiny Clean}
		\centerline{\tiny PSNR}
		\end{minipage}
		\vspace{0.03in}
		\caption{The difference maps (red boxes shown at the top-right) between the input rain image and results by deraining methods that suffer a PSNR drop. (Brighter indicates a higher difference.) We can see that \cite{luo:iccv:2015:dsc,yang:cvpr:2017:j,zhang:cvpr:2018:did,li:eccv:2018:rsecan} tend to over-derain the image.}
		\label{fig:example3}

\end{figure}

We also show the performance of non-CNN-based state-of-the-art methods in Table~\ref{table:benchmark}.
We have an interesting observation here that the input rain images have similar or even higher average PSNR and SSIM scores compared with those of the derained results by the state-of-the-art derainers. As demonstrated in Figure~\ref{fig:example3}, it is mainly caused by over deraining. Even though \cite{luo:iccv:2015:dsc} is less dependent on training data (but still depends on a learned dictionary) as the deep learning methods (\cite{yang:cvpr:2017:j,zhang:cvpr:2018:did,li:eccv:2018:rsecan}), it fails when the rain exhibits unseen appearances and mistakenly removes the structures that are similar to rain streaks.

\begin{figure*}
\centering
\begin{minipage}[t]{0.85\textwidth}
\centering
\includegraphics[width=\textwidth]{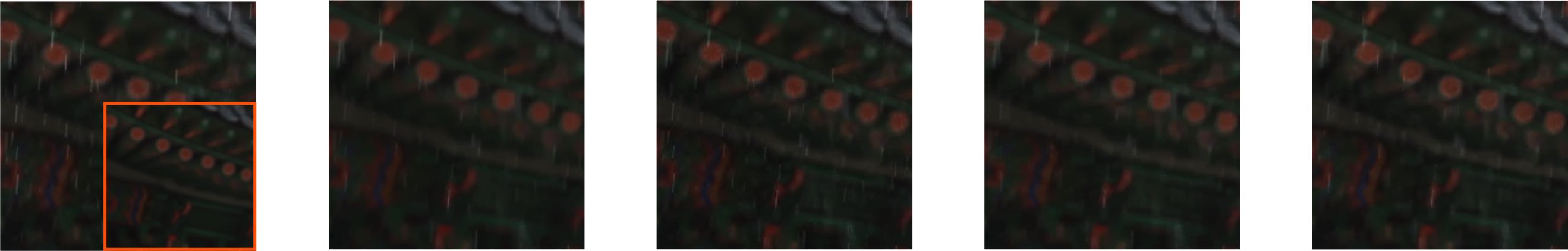}
\end{minipage}

\begin{minipage}[t]{0.17\textwidth}
\centerline{\footnotesize (a) Rain / Clean Image}
\centerline{\footnotesize 31.06 / 0.9108 }
\end{minipage}
\centering
\begin{minipage}[t]{0.17\linewidth}
\centerline{\footnotesize (b) DSC~\cite{luo:iccv:2015:dsc}}
\centerline{\footnotesize 34.49 / 0.9316}
\end{minipage}
\begin{minipage}[t]{0.17\linewidth}
\centerline{\footnotesize (c) LP~\cite{li:cvpr:2016:lp}}
\centerline{\footnotesize 34.42 / 0.9488 }
\end{minipage}
\begin{minipage}[t]{0.17\linewidth}
\centerline{\footnotesize (d) SILS~\cite{gu:iccv:2017:sils}}
\centerline{\footnotesize 33.20 / 0.9463 }
\end{minipage}
\begin{minipage}[t]{0.17\linewidth}
\centerline{\footnotesize (e) Clearing~\cite{fu:tip:2016:clearing}}
\centerline{\footnotesize 31.82 / 0.9353}
\end{minipage}

\begin{minipage}[t]{\textwidth}
\centerline{  }
\end{minipage}

\begin{minipage}[t]{0.85\textwidth}
\includegraphics[width=\textwidth]{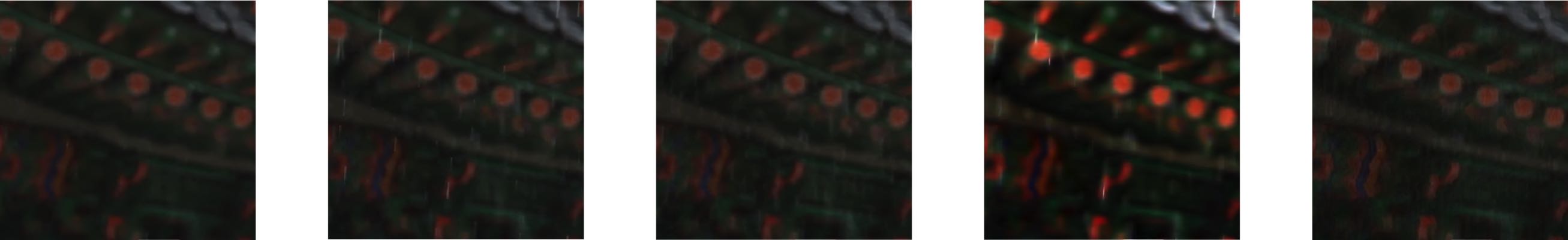}
\end{minipage}

\begin{minipage}[t]{0.17\linewidth}
\centerline{\footnotesize \textbf{(f)} \textbf{Our SPANet}}
\centerline{\footnotesize \textbf{38.22} / \textbf{0.9764}}
\end{minipage}
\centering
\begin{minipage}[t]{0.17\textwidth}
\centerline{\footnotesize (g) DDN~\cite{fu:cvpe:2017:ddn}}
\centerline{\footnotesize 33.94 / 0.9460 }
\end{minipage}
\centering
\begin{minipage}[t]{0.17\textwidth}
\centerline{\footnotesize (h) JORDER~\cite{yang:cvpr:2017:j}}
\centerline{\footnotesize 35.09 / 0.9495}
\end{minipage}
\centering
\begin{minipage}[t]{0.17\textwidth}
\centerline{\footnotesize (i) DID-MDN~\cite{zhang:cvpr:2018:did}}
\centerline{\footnotesize 21.69 / 0.8018}
\end{minipage}
\begin{minipage}[t]{0.17\textwidth}
\centerline{\footnotesize (j) RESCAN~\cite{li:eccv:2018:rsecan}}
\centerline{\footnotesize 34.35 / 0.9265}
\end{minipage}
\vspace{1.5mm}
\caption{Visual comparison of SPANet with the state-of-the-art derainers. PSNR/SSIM results are included for reference.}
\label{fig:bench}
\vspace{-4mm}
\end{figure*}

\begin{figure*}[t]
\centering
\begin{minipage}[t]{0.15\linewidth}
\includegraphics[width=\textwidth]{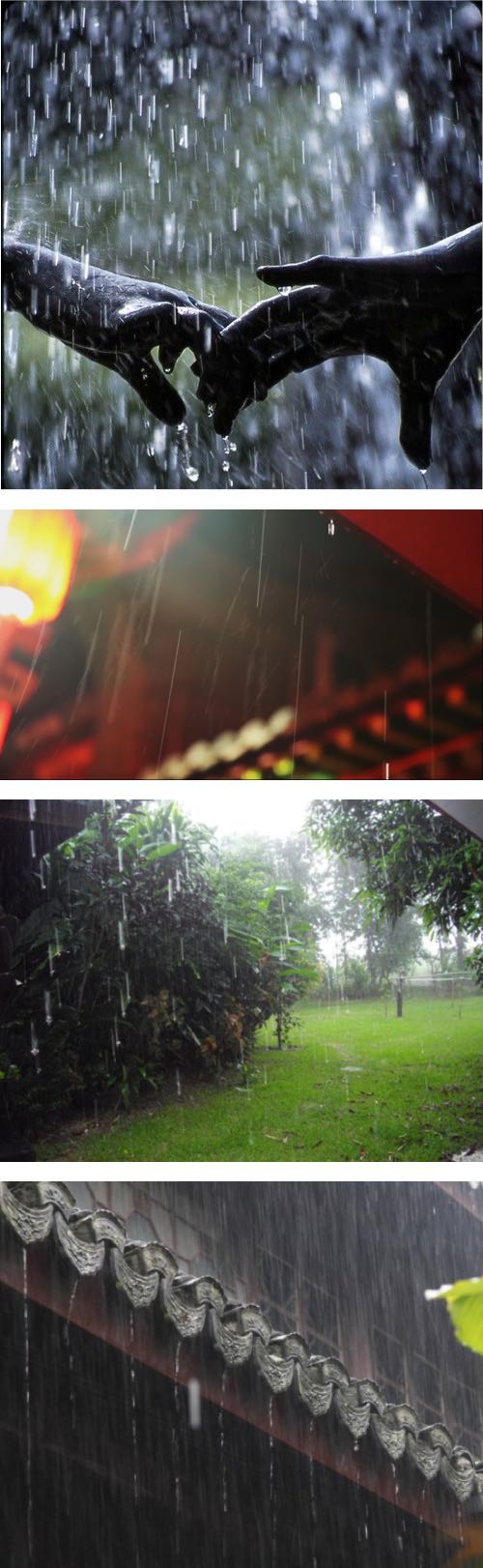}
\centerline{\footnotesize (a) Rain}
\end{minipage}
\begin{minipage}[t]{0.15\linewidth}
\includegraphics[width=\textwidth]{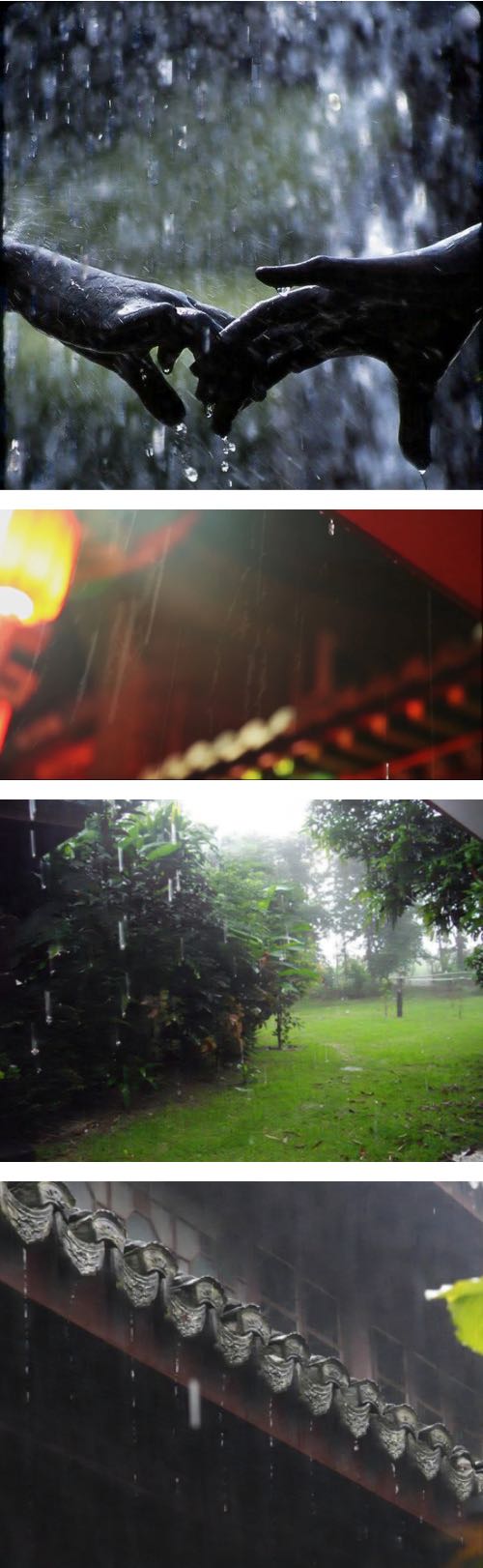}
\centerline{\footnotesize (b) DDN~\cite{fu:cvpe:2017:ddn}}
\end{minipage}
\begin{minipage}[t]{0.15\linewidth}
\includegraphics[width=\textwidth]{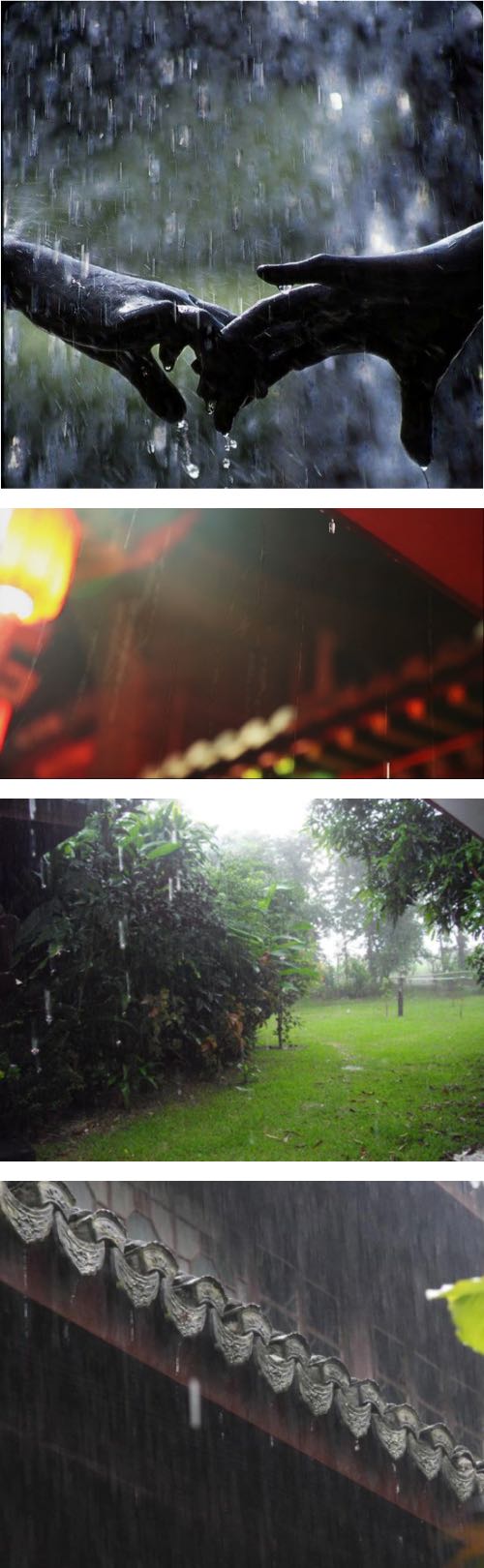}
\centerline{\footnotesize (c) JORDER~\cite{yang:cvpr:2017:j}}
\end{minipage}
\begin{minipage}[t]{0.15\linewidth}
\includegraphics[width=\textwidth]{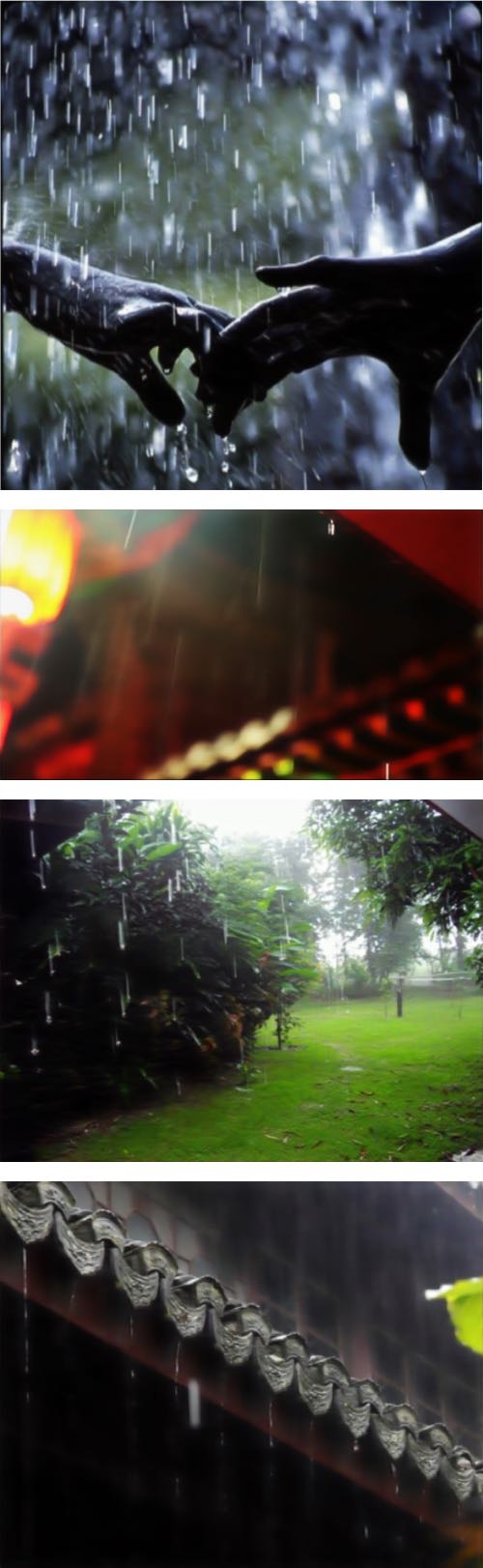}
\centerline{\footnotesize (d) DID-MDN~\cite{zhang:cvpr:2018:did}}
\end{minipage}
\begin{minipage}[t]{0.15\linewidth}
\includegraphics[width=\textwidth]{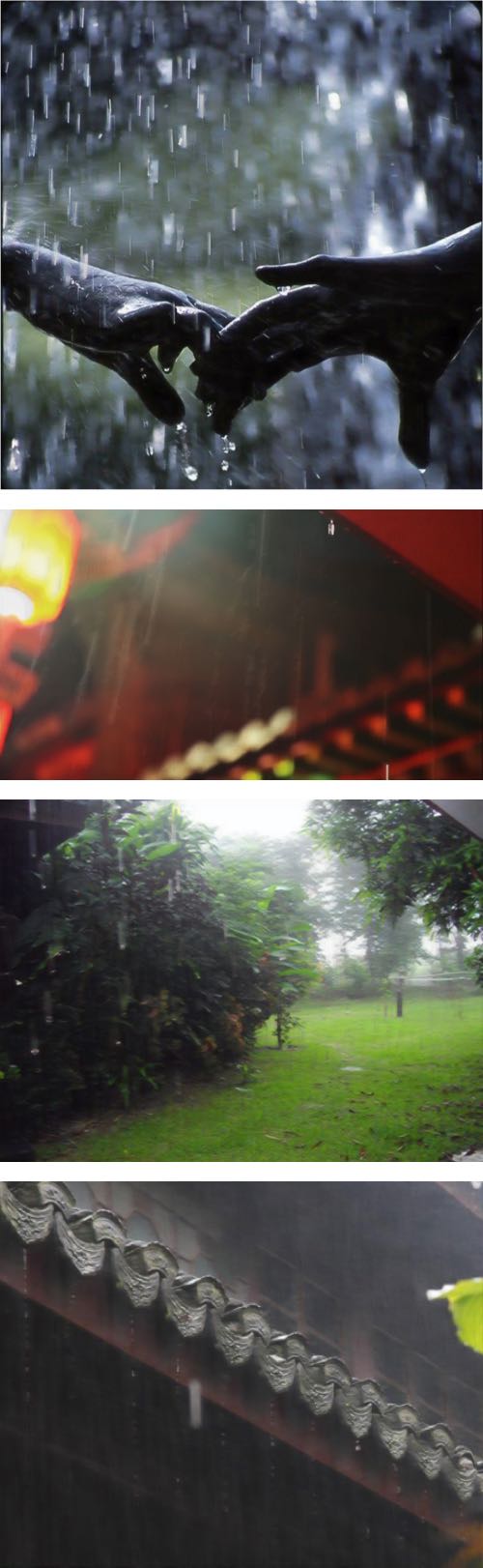}
\centerline{\footnotesize (e) RESCAN~\cite{li:eccv:2018:rsecan}}
\end{minipage}
\begin{minipage}[t]{0.15\linewidth}
\includegraphics[width=\textwidth]{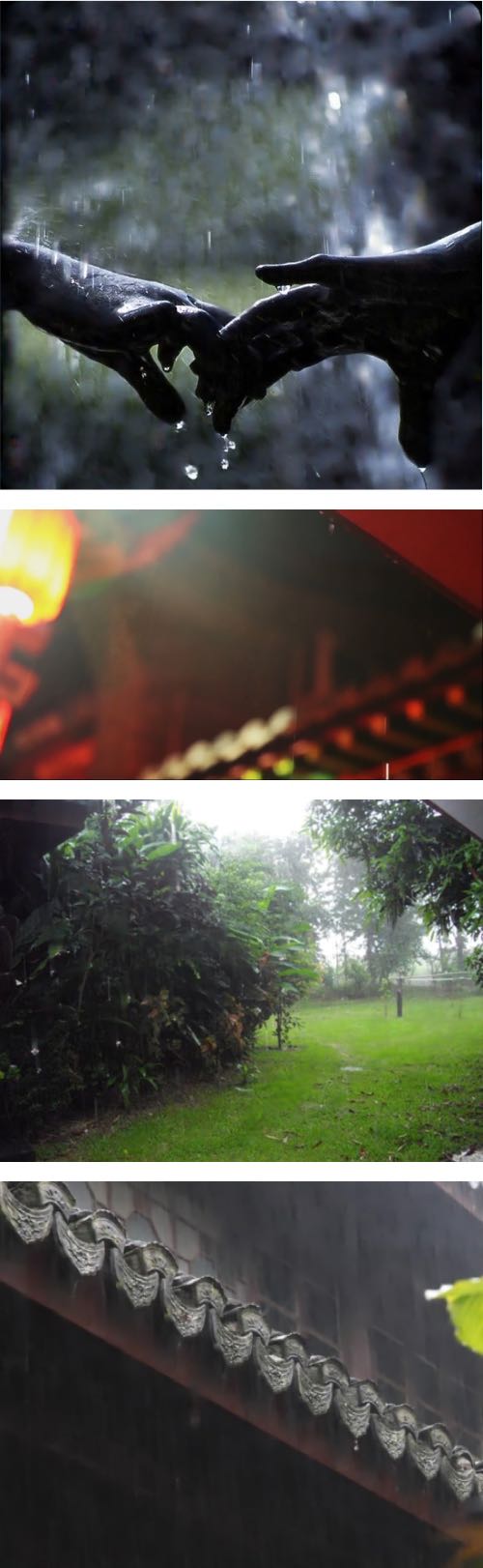}
\centerline{\footnotesize (f) \textbf{Our SPANet}}
\end{minipage}
\vspace{0.05in}
\caption{Visual comparison of SPANet with the state-of-the-art CNN-based derainers on some real rain images collected from previous derain papers and from the Internet.}
\label{fig:real}
\end{figure*}

{\bf Evaluation on the proposed SPANet.} Table~\ref{table:benchmark} reports the performance of our SPANet, trained on the proposed dataset. It achieves a superior deraining performance compared to the state-of-the-art derainers. This is because SPANet can identify the rain streak regions and remove them accurately. Figure~\ref{fig:bench} shows a visual example from our test set. We can see that while methods (b)$-$(e) tend to leave rain streaks unremoved and methods (g)$-$(j) tend to corrupt the background, the proposed SPANet (f) can produce much cleaner result. We also show some deraining examples on rain images collected from previous derain papers and the Internet in Figure~\ref{fig:real}. While existing derainers fail to remove the rain streaks and some of them tend to darken or blur the background, our SPANet can handle different kinds of rain streaks and preserve more details. Table~\ref{table:syn} compares the performances of SPANet with the state-of-the-art derainers on the synthetic test set from~\cite{zhang:cvpr:2018:did}, demonstrating the effectiveness of SPANet.

\begin{table*}[t]
\vspace{-2mm}
\centering
\resizebox{\textwidth}{!}{%
\begin{tabular}{@{}cccccccccc@{}}
\toprule
Methods & Input & \begin{tabular}[c]{@{}c@{}}DSC \cite{luo:iccv:2015:dsc}\end{tabular} & \begin{tabular}[c]{@{}c@{}}LP \cite{li:cvpr:2016:lp} \end{tabular} & \begin{tabular}[c]{@{}c@{}}Clear\cite{fu:tip:2016:clearing}\end{tabular} & \begin{tabular}[c]{@{}c@{}}JORDER \cite{yang:cvpr:2017:j}\end{tabular} & \begin{tabular}[c]{@{}c@{}}DDN \cite{fu:cvpe:2017:ddn}\end{tabular} & \begin{tabular}[c]{@{}c@{}}JBO\cite{zhu:iccv:2017:jbo}\end{tabular} & \begin{tabular}[c]{@{}c@{}}DID-MDN\cite{zhang:cvpr:2018:did}\end{tabular} & \textbf{Our SPANet} \\ \midrule
DID-MDN Test Set & 0.7781/21.15 & 0.7896/21.44 & 0.8352/22.75 & 0.8422/22.07 & 0.8622/24.32 & 0.8978/ 27.33 & 0.8522/23.05 & 0.9087/ 27.95 & \textbf{0.9342/30.05} \\ \bottomrule
\end{tabular}%
}
\caption{Comparison on the test set from \cite{zhang:cvpr:2018:did}. SPANet is trained on the synthetic dataset from~\cite{zhang:cvpr:2018:did}.}
\label{table:syn}
\vspace{-3mm}
\end{table*}

{\bf Internal analysis.} We verify the importance of the spatial attentive module (SAM) and different ways of using it in Table~\ref{table:ablation}. $B_a$ is a basic Resnet-like network that does not use SAM. $B_b$, $B_c$, and $B_f$ represent three variants of using only one SAM for four times (recall that we have four SAB blocks), four SAMs, and four SAMs that share the same weights for all operations, respectively. While we can see that all variants of incorporating the SAM improve the performance, $B_f$ performs the best, as sharing the weights makes the deraining process inter-dependent on the four SAB blocks, which allows more attention to be put to the challenging real rain streak distributions.
$B_d$ is the SPANet but without the above attention branch in SAM. The comparison between $B_d$ and $B_f$ shows that attention branch is effective in leveraging the local contextual information aggregated from different directions.
$B_e$ is a variant that removes the attention loss supervision. It demonstrates the importance of providing explicit supervision on the attention map generation process.

\begin{table}[h]

\resizebox{\linewidth}{!}{%
\begin{tabular}{@{}ccccccc@{}}
\toprule
Methods & \multicolumn{1}{|c}{$B_a$} & $B_b$ & $B_c$ & $B_d$ & $B_e$ & $B_f$ \\ \midrule
Resnet & \multicolumn{1}{|c}{\checkmark} & \checkmark & \checkmark & \checkmark & \checkmark & \checkmark \\
Single SAM & \multicolumn{1}{|c}{} & \checkmark &  &  &  &  \\
4 SAMs w/o shared weights & \multicolumn{1}{|c}{} &  & \checkmark &  &  &  \\
4 SAMs w/ shared weights & \multicolumn{1}{|c}{} &  &  & \checkmark & \checkmark & \checkmark \\
Self-Attention branch & \multicolumn{1}{|c}{} & \checkmark & \checkmark &  & \checkmark & \checkmark \\
Attention Loss & \multicolumn{1}{|c}{} & \checkmark & \checkmark & \checkmark &  & \checkmark \\ \midrule
PSNR & \multicolumn{1}{|c}{37.43} & 37.43 & 37.47 & 37.70 & 37.39 & {\bf 38.06} \\
SSIM & \multicolumn{1}{|c}{0.9856} & 0.9854 & 0.9854 & 0.9858 & 0.9856 & {\bf 0.9867} \\ \bottomrule
\end{tabular}%
}
\caption{Internal analysis of the proposed SPANet. The best performance is marked in {\bf bold}. }
\label{table:ablation}
\vspace{-2mm}
\end{table}

\begin{figure}[h]
	\centering
	\vspace{-2mm}
		\subfigure[Input]{
			\includegraphics[width = 0.22\linewidth]{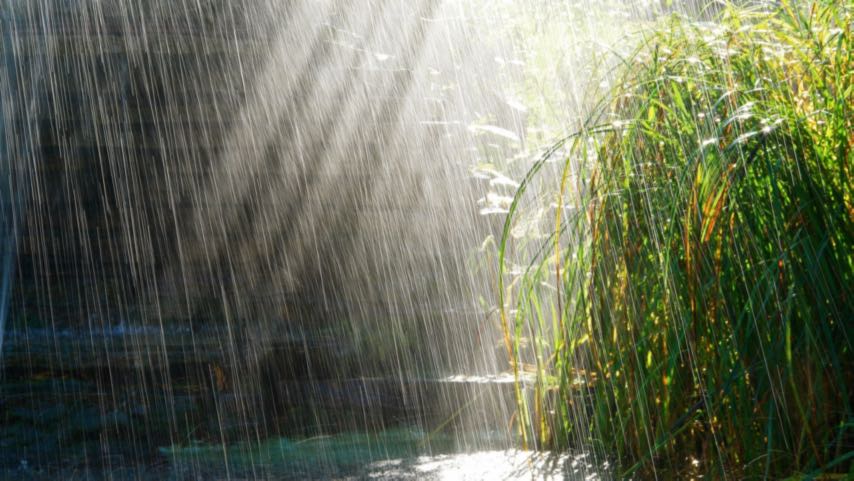}
		}
		\subfigure[JORDER]{
			\includegraphics[width = 0.22\linewidth]{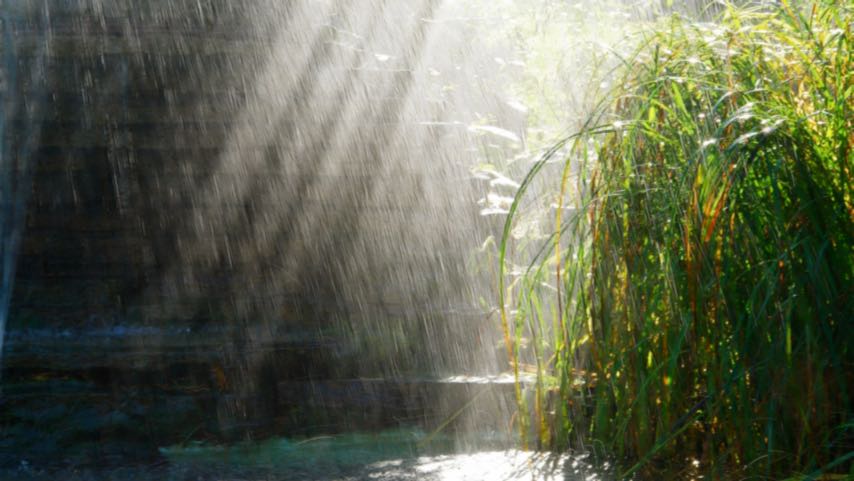}
		}
		\subfigure[DID-MDN]{
			\includegraphics[width = 0.22\linewidth]{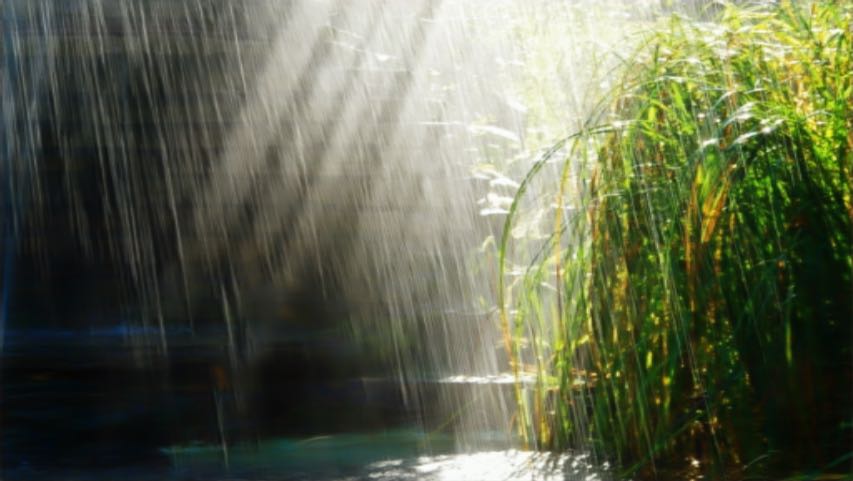}
		}
		\subfigure[Our SPANet]{
			\includegraphics[width = 0.22\linewidth]{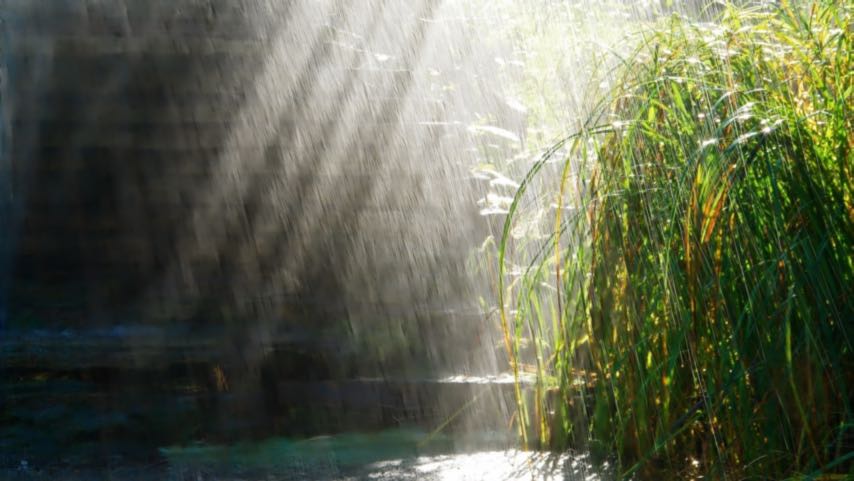}
		}

		\caption{Failure case. Our method fails to remove extremely dense rain streaks.}
		\label{fig:example2}
		\vspace{-4mm}
\end{figure}

\section{Conclusion and Future Work}
In this paper, we have presented a method to produce a high-quality clean image from a sequence of real rain images, by considering temporal priors together with human supervision. Based on this method, we have constructed a large-scale dataset of $\sim$$29.5K$ rain/clean image pairs that cover a wide range of natural rain scenes. Experiments show that the performances of state-of-the-art CNN-based derainers can be significantly improved by training on the proposed dataset. We have also benchmarked state-of-the-art derainers on the proposed test set. We find that the stochastic distributions of real rain streaks, especially the varying appearances of rain streaks, often fail these methods.
To this end, we present a novel spatial attentive network (SPANet) that can learn to identify and remove rain streaks in a local-to-global spatial attentive manner. Extensive evaluations demonstrate the superiority of the proposed method over the state-of-the-art derainers.

Our method does have limitations. One example is given in Figure~\ref{fig:example2}, which shows that our method fails when processing haze-like heavy rain. It is because the proposed dataset generation method fails to select clean pixels from the misty video frames. As a result, the proposed network produces a haze-like result.

Currently, our dataset generation method relies on human judgements. This is partly due to the fact that there are no existing metrics that can assess the generated rain-free images, without clean images for reference. It would be interesting to develop an unsupervised mechanism for this purpose in the future.

{\bf Acknowledgement.} This work was supported by NSFC (\#91748104, \#U1811463, \#61632006, \#61425002, \#61751203), Key Research and Development Program of China (\#2018YFC0910506), and the Open Project Program of the State Key Lab of CAD\&CG (\#A1901).

{\small
\bibliographystyle{ieee}
\bibliography{reference}

\begin{thebibliography}{10}\itemsep=-1pt

\bibitem{abdelhamed:cvpr:2018:dndata}
Abdelrahman Abdelhamed, Stephen Lin, and Michael Brown.
\newblock A high-quality denoising dataset for smartphone cameras.
\newblock In {\em CVPR}, 2018.

\bibitem{anaya:arxiv:2014:RENOIR}
Josue Anaya and Adrian Barbu.
\newblock Renoir - a benchmark dataset for real noise reduction evaluation.
\newblock {\em arXiv:1409.8230}, 2014.

\bibitem{bell:cvpr:2016:ion}
Sean Bell, C. Lawrence~Zitnick, Kavita Bala, and Ross Girshick.
\newblock Inside-outside net: Detecting objects in context with skip pooling
  and recurrent neural networks.
\newblock In {\em CVPR}, 2016.

\bibitem{bossu:ijcv:2011:raindetection}
J{\'e}r{\'e}mie Bossu, Nicolas Hauti{\`e}re, and Jean-Philippe Tarel.
\newblock Rain or snow detection in image sequences through use of a histogram
  of orientation of streaks.
\newblock {\em IJCV}, 2011.

\bibitem{chang:iccv:2017:lpnr}
Yi Chang, Luxin Yan, and Sheng Zhong.
\newblock Transformed low-rank model for line pattern noise removal.
\newblock In {\em ICCV}, 2017.

\bibitem{chen:tip:2014:rpra}
Jie Chen and Lap{-}Pui Chau.
\newblock A rain pixel recovery algorithm for videos with highly dynamic
  scenes.
\newblock {\em IEEE TIP}, 2014.

\bibitem{chen:cvpr:2018:videoderaincnn}
Jie Chen, Cheen-Hau Tan, Junhui Hou, Lap-Pui Chau, and He Li.
\newblock Robust video content alignment and compensation for rain removal in a
  cnn framework.
\newblock In {\em CVPR}, 2018.

\bibitem{chen:iccv:2013:generalized}
Yi~Lei Chen and Chiou~Ting Hsu.
\newblock A generalized low-rank appearance model for spatio-temporally
  correlated rain streaks.
\newblock In {\em ICCV}, 2013.

\bibitem{du:pr:2018:grad}
Shuangli Du, Yiguang Liu, Mao Ye, Zhenyu Xu, Jie Li, and Jianguo Liu.
\newblock Single image deraining via decorrelating the rain streaks and
  background scene in gradient domain.
\newblock {\em PR}, 2018.

\bibitem{fu:tip:2016:clearing}
Xueyang Fu, Jiabin Huang, Xinghao Ding, Yinghao Liao, and John Paisley.
\newblock Clearing the skies: A deep network architecture for single-image rain
  streaks removal.
\newblock {\em IEEE TIP}, 2017.

\bibitem{fu:cvpe:2017:ddn}
Xueyang Fu, Jiabin Huang, Delu Zeng, Yue Huang, Xinghao Ding, and John Paisley.
\newblock Removing rain from single images via a deep detail network.
\newblock In {\em CVPR}, 2017.

\bibitem{garg:cvpr:2004:detection}
Kshitiz Garg and Shree~K. Nayar.
\newblock Detection and removal of rain from videos.
\newblock In {\em CVPR}, 2004.

\bibitem{garg:ijcv:2007:vision}
Kshitiz Garg and Shree~K Nayar.
\newblock Vision and rain.
\newblock {\em IJCV}, 2007.

\bibitem{gu:iccv:2017:sils}
Shuhang Gu, Deyu Meng, Wangmeng Zuo, and Lei Zhang.
\newblock Joint convolutional analysis and synthesis sparse representation for
  single image layer separation.
\newblock In {\em ICCV}, 2017.

\bibitem{li:acmmm:2018:nled}
Wei Zhang Huiyou Chang Le Dong Liang~Lin Guanbin~Li, Xiang~He.
\newblock Non-locally enhanced encoder-decoder network for single image
  de-raining.
\newblock In {\em ACM MM}, 2018.

\bibitem{he:cvpr:2016:resnet}
Kaiming He, Xiangyu Zhang, Shaoqing Ren, and Jian Sun.
\newblock Deep residual learning for image recognition.
\newblock In {\em CVPR}, 2016.

\bibitem{hu:cvpr:2018:senet}
Jie Hu, Li Shen, and Gang Sun.
\newblock Squeeze-and-excitation networks.
\newblock In {\em CVPR}, 2018.

\bibitem{hu:cvpr:2018:dsc}
Xiaowei Hu, Lei Zhu, Chi-Wing Fu, Jing Qin, and Pheng-Ann Heng.
\newblock Direction-aware spatial context features for shadow detection.
\newblock In {\em CVPR}, 2018.

\bibitem{jiang:cvpr:2017:dip}
Tai-Xiang Jiang, Ting-Zhu Huang, Xi-Le Zhao, Liang-Jian Deng, and Yao Wang.
\newblock A novel tensor-based video rain streaks removal approach via
  utilizing discriminatively intrinsic priors.
\newblock In {\em CVPR}, 2017.

\bibitem{kang:tip:2012:imgdecomp}
Li-Wei Kang, Chia-Wen Lin, and Yu-Hsiang Fu.
\newblock Automatic single-image-based rain streaks removal via image
  decomposition.
\newblock {\em IEEE TIP}, 2012.

\bibitem{kim:tip:2015:videoderain}
Jin-Hwan Kim, Jae-Young Sim, and Chang-Su Kim.
\newblock Video deraining and desnowing using temporal correlation and low-rank
  matrix completion.
\newblock {\em IEEE TIP}, 2015.

\bibitem{kingma:iclr:2014:adam}
Diederik~P. Kingma and Jimmy Ba.
\newblock Adam: A method for stochastic optimization.
\newblock In {\em ICLR}, 2015.

\bibitem{le:arxiv:2015:simple}
Quoc Le, Navdeep Jaitly, and Geoffrey Hinton.
\newblock A simple way to initialize recurrent networks of rectified linear
  units.
\newblock {\em arXiv:1504.00941}, 2015.

\bibitem{li:cvpr:2018:mcsc}
Minghan Li, Qi Xie, Qian Zhao, Wei Wei, Shuhang Gu, Jing Tao, and Deyu Meng.
\newblock Video rain streak removal by multiscale convolutional sparse coding.
\newblock In {\em CVPR}, 2018.

\bibitem{li:eccv:2018:rsecan}
Xia Li, Jianlong Wu, Zhouchen Lin, Hong Liu, and Hongbin Zha.
\newblock Recurrent squeeze-and-excitation context aggregation net for single
  image deraining.
\newblock In {\em ECCV}, 2018.

\bibitem{li:cvpr:2016:lp}
Yu Li, Robby~T Tan, Xiaojie Guo, Jiangbo Lu, and Michael Brown.
\newblock Rain streak removal using layer priors.
\newblock In {\em CVPR}, 2016.

\bibitem{liu:cvpr:2018:j4r}
Jiaying Liu, Wenhan Yang, Shuai Yang, and Zongming Guo.
\newblock Erase or fill? deep joint recurrent rain removal and reconstruction
  in videos.
\newblock In {\em CVPR}, 2018.

\bibitem{liu:cis:2009:pixel}
Peng Liu, Jing Xu, Jiafeng Liu, and Xianglong Tang.
\newblock {Pixel Based Temporal Analysis Using Chromatic Property for Removing
  Rain from Videos.}
\newblock {\em CIS}, 2009.

\bibitem{luo:iccv:2015:dsc}
Yu Luo, Yong Xu, and Hui Ji.
\newblock Removing rain from a single image via discriminative sparse coding.
\newblock In {\em ICCV}, 2015.

\bibitem{mao:cvpr:2017:pd}
Jiayuan Mao, Tete Xiao, Yuning Jiang, and Zhimin Cao.
\newblock What can help pedestrian detection?
\newblock In {\em CVPR}, 2017.

\bibitem{rainstreaks}
motionvfx.
\newblock https://www.motionvfx.com/mplugs-48.html, 2014.

\bibitem{nam:cvpr:2016:hadenoise}
Seonghyeon Nam, Youngbae Hwang, Yasuyuki Matsushita, and Seon Joo~Kim.
\newblock A holistic approach to cross-channel image noise modeling and its
  application to image denoising.
\newblock In {\em CVPR}, 2016.

\bibitem{plotz:cvpr:2017:noisebench}
Tobias Plotz and Stefan Roth.
\newblock Benchmarking denoising algorithms with real photographs.
\newblock In {\em CVPR}, 2017.

\bibitem{pytorch}
PyTorch.
\newblock http://pytorch.org.

\bibitem{ren:cvpr:2017:md}
Weihong Ren, Jiandong Tian, Zhi Han, Antoni Chan, and Yandong Tang.
\newblock Video desnowing and deraining based on matrix decomposition.
\newblock In {\em CVPR}, 2017.

\bibitem{santhaseelan:ijcv:2015:utilizing}
Varun Santhaseelan and Vijayan~K Asari.
\newblock Utilizing local phase information to remove rain from video.
\newblock {\em IJCV}, 2015.

\bibitem{song:cvpr:2018:vital}
Yibing Song, Chao Ma, Xiaohe Wu, Lijun Gong, Linchao Bao, Wangmeng Zuo, Chunhua
  Shen, Rynson~W.H. Lau, and Ming-Hsuan Yang.
\newblock Vital: Visual tracking via adversarial learning.
\newblock In {\em CVPR}, 2018.

\bibitem{wang:tip:2004:ssim}
Zhou Wang, Alan~C Bovik, Hamid~R Sheikh, and Eero~P Simoncelli.
\newblock {Image quality assessment - from error visibility to structural
  similarity}.
\newblock {\em IEEE TIP}, 2004.

\bibitem{wei:iccv:2017:ds}
Wei Wei, Lixuan Yi, Qi Xie, Qian Zhao, Deyu Meng, and Zongben Xu.
\newblock Should we encode rain streaks in video as deterministic or
  stochastic?
\newblock In {\em ICCV}, 2017.

\bibitem{yang:cvpr:2017:j}
Wenhan Yang, Robby~T Tan, Jiashi Feng, Jiaying Liu, Zongming Guo, and Shuicheng
  Yan.
\newblock Deep joint rain detection and removal from a single image.
\newblock In {\em CVPR}, 2017.

\bibitem{yu:iclr:2015:dilated}
Fisher Yu and Vladlen Koltun.
\newblock Multi-scale context aggregation by dilated convolutions.
\newblock In {\em ICLR}, 2016.

\bibitem{zhang:cvpr:2018:did}
He Zhang and Vishal~M. Patel.
\newblock Density-aware single image de-raining using a multi-stream dense
  network.
\newblock In {\em CVPR}, 2018.

\bibitem{zhang:arxiv:2017:gan}
He Zhang, Vishwanath Sindagi, and Vishal Patel.
\newblock Image de-raining using a conditional generative adversarial network.
\newblock {\em arXiv:1701.05957}, 2017.

\bibitem{zhang:icme:2006:temporal}
Xiaopeng Zhang, Hao Li, Yingyi Qi, Wee~Kheng Leow, and Teck~Khim Ng.
\newblock Rain removal in video by combining temporal and chromatic properties.
\newblock In {\em ICME}, 2006.

\bibitem{fan:acmmm:2018:rgffn}
Xueyang Fu Yue Huang Xinghao~Ding Zhiwen~Fan, Huafeng~Wu.
\newblock Residual-guide feature fusion network for single image deraining.
\newblock In {\em ACM MM}, 2018.

\bibitem{zhu:cvpr:2016:denoise}
Fengyuan Zhu, Guangyong Chen, and Pheng-Ann Heng.
\newblock From noise modeling to blind image denoising.
\newblock In {\em CVPR}, 2016.

\bibitem{zhu:iccv:2017:jbo}
Lei Zhu, Chi-Wing Fu, Dani Lischinski, and Pheng-Ann Heng.
\newblock Joint bi-layer optimization for single-image rain streak removal.
\newblock In {\em ICCV}, 2017.

\bibitem{zhu:cvpr:2016:tsd}
Zhe Zhu, Dun Liang, Songhai Zhang, Xiaolei Huang, Baoli Li, and Shimin Hu.
\newblock Traffic-sign detection and classification in the wild.
\newblock In {\em CVPR}, 2016.

\end{thebibliography}
}

\end{document}